\documentclass[10pt,twocolumn,letterpaper]{article}

\usepackage[pagenumbers]{iccv} %
\usepackage{multirow}
\usepackage{booktabs}
\usepackage{graphicx}    %
\usepackage{adjustbox}    %
\usepackage{threeparttable}
\usepackage[T1]{fontenc}
\usepackage{pifont}
\usepackage{amsmath}
\usepackage{bbm}
\usepackage{colortbl}
\usepackage{color}
\usepackage[T1]{fontenc}
\usepackage{textcomp}
\usepackage{stfloats}
\usepackage{caption}
\usepackage{microtype}
\usepackage{comment}

\usepackage{arydshln}

\usepackage{microtype}

\newcommand{\red}[1]{{\color{red}#1}}

\newcommand{\probP}{\text{I\kern-0.15em P}}

\definecolor{iccvblue}{rgb}{0.21,0.49,0.74}
\usepackage[pagebackref,breaklinks,colorlinks,allcolors=iccvblue]{hyperref}
\def\modelName{[$\mathrm{OURMODEL}$]}
\newcommand{\keypoint}[1]{\vspace{0.1cm}\noindent\textbf{#1}\quad}
\def\etal{\emph{et al}.}
\def\modelName{\textsc{UniCoRN}}
\def\dataName{\textsc{MetaRestore}}
\definecolor{LightCyan}{rgb}{0.95,0.95,0.95}
\newcolumntype{a}{>{\columncolor{LightCyan}}c}

\title{\textsc{UniCoRN}: Latent Diffusion-based \underline{Uni}fied \underline{Co}ntrollable Image \underline{R}estoration \underline{N}etwork across Multiple Degradations}

\author{Debabrata Mandal\textsuperscript{} \hspace{.1cm} Soumitri Chattopadhyay\textsuperscript{} \hspace{.1cm} Guansen Tong\textsuperscript{} \hspace{.1cm} Praneeth Chakravarthula\textsuperscript{} \\
\textsuperscript{}University of North Carolina at Chapel Hill, USA\\
{\tt\small \{debman, soumitri, gtong, cpk\}@cs.unc.edu} \\
}

\begin{document}

\twocolumn[{%
  \renewcommand\twocolumn[1][]{#1}%
  \maketitle
  \vspace{-1.1cm}
  \begin{center}
   \centering
   \includegraphics[width=\textwidth]{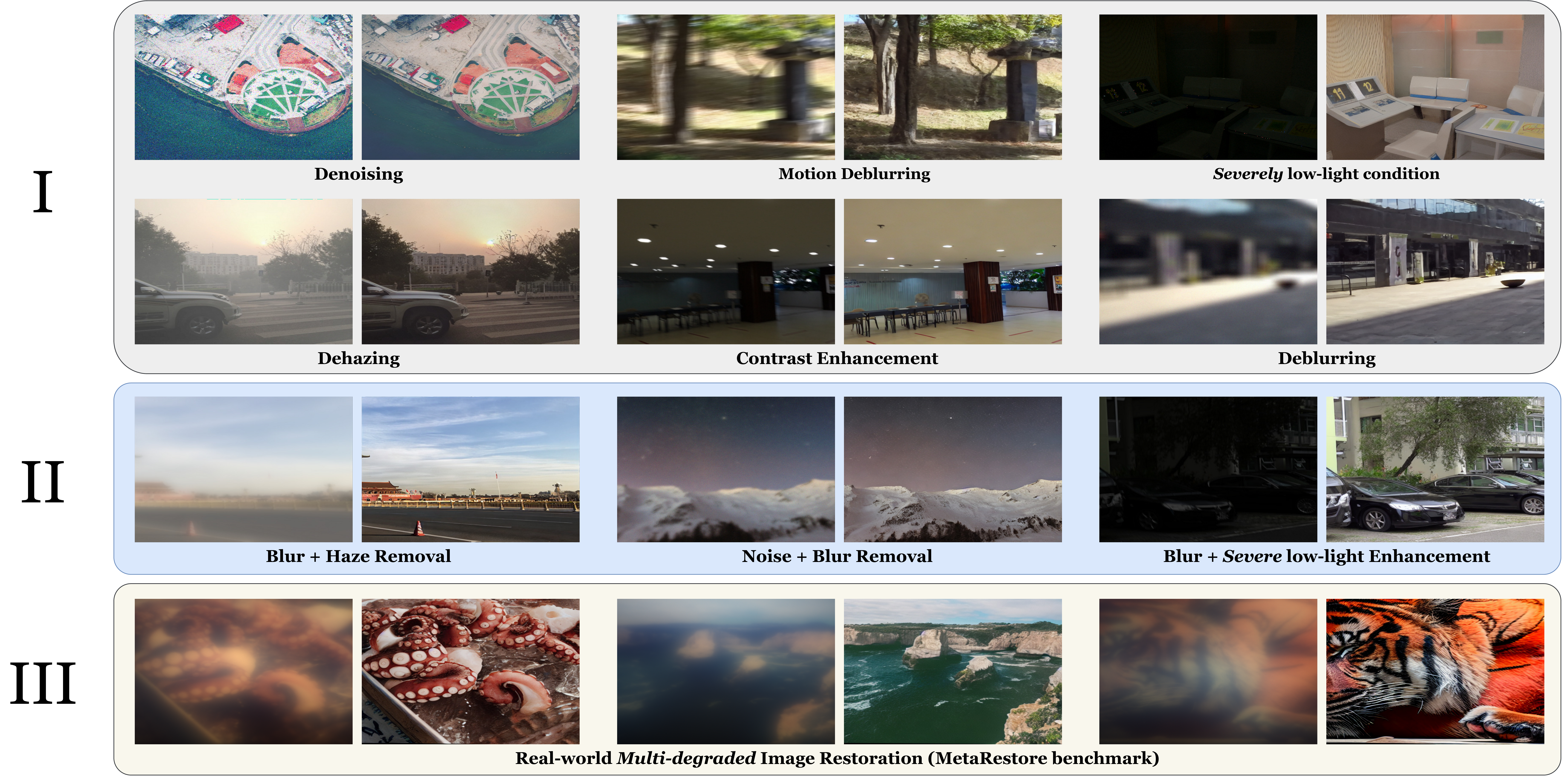}
   \captionof{figure}{
We present \modelName, a unified image restoration approach for handling multiple degradations simultaneously. Our approach shows robust performance on existing benchmarks and excels on \dataName, exhibiting multiple degradations. \modelName, based on stable diffusion, is corruption-agnostic, flexible, and scalable.}
   \label{fig1}
  \end{center}%
 }]

\begin{abstract}

Image restoration is essential for enhancing degraded images across computer vision tasks. However, most existing methods address only a single type of degradation (e.g., blur, noise, or haze) at a time, limiting their real-world applicability where multiple degradations often occur simultaneously.
In this paper, we propose {\modelName},  a unified image restoration approach capable of handling multiple degradation types simultaneously using a multi-head diffusion model. Specifically, we uncover the potential of low-level visual cues extracted from images in guiding a controllable diffusion model for real-world image restoration and we design a multi-head control network adaptable via a mixture-of-experts strategy. We train our model without any prior assumption of specific degradations, through a smartly designed curriculum learning recipe. Additionally, we also introduce {\dataName}, a metalens imaging benchmark containing images with multiple degradations and artifacts.
Extensive evaluations on several challenging datasets, including our benchmark, demonstrate that our method achieves significant performance gains and can robustly restore images with severe degradations.

\noindent Project page: \href{https://codejaeger.github.io/unicorn-gh}{https://codejaeger.github.io/unicorn-gh}

\end{abstract}
    
\def\red#1{\textcolor[rgb]{1,0,0}{#1}}
\def\blue#1{\textcolor[rgb]{0,0,1}{#1}}
\def\black#1{\textcolor[rgb]{0,0,0}{#1}}
\def\orange#1{\textcolor[rgb]{1,0.49,0}{#1}}
\def\yellow#1{\textcolor[rgv]{1,1,0}{#1}}
\def\etal{\emph{et al}.}

\section{Introduction}
\label{sec:intro}

Image restoration \cite{su2022survey, ali2023vision, zhai2023comprehensive, zamir2022restormer, potlapalli2024promptir} is a fundamental ill-posed inverse problem \cite{ali2023vision, zamir2022restormer, potlapalli2024promptir} in computer vision aimed at recovering high-quality images from its degraded, lower-quality counterparts. These degradations often arise during image acquisition due to undesirable environmental conditions \cite{lee2016review, liu2021benchmarking} (e.g., low-light, haze, rain, blur) or inherent camera limitations \cite{zhai2023comprehensive, wang2023effect, fergus2006removing} (e.g., sensor noise, chromatic aberrations), and can impact downstream applications. 
Most image restoration approaches have focused on addressing a single type of degradation, such as denoising \cite{fan2019brief}, dehazing \cite{lee2016review, song2023dehazeformer}, deblurring \cite{cho2021rethinking, zhang2022deep}, deraining \cite{zamir2022restormer, li2020all}, or low-light enhancement \cite{liu2021benchmarking, park2017low}.

\begin{figure}[t]
    \includegraphics[width=\columnwidth]{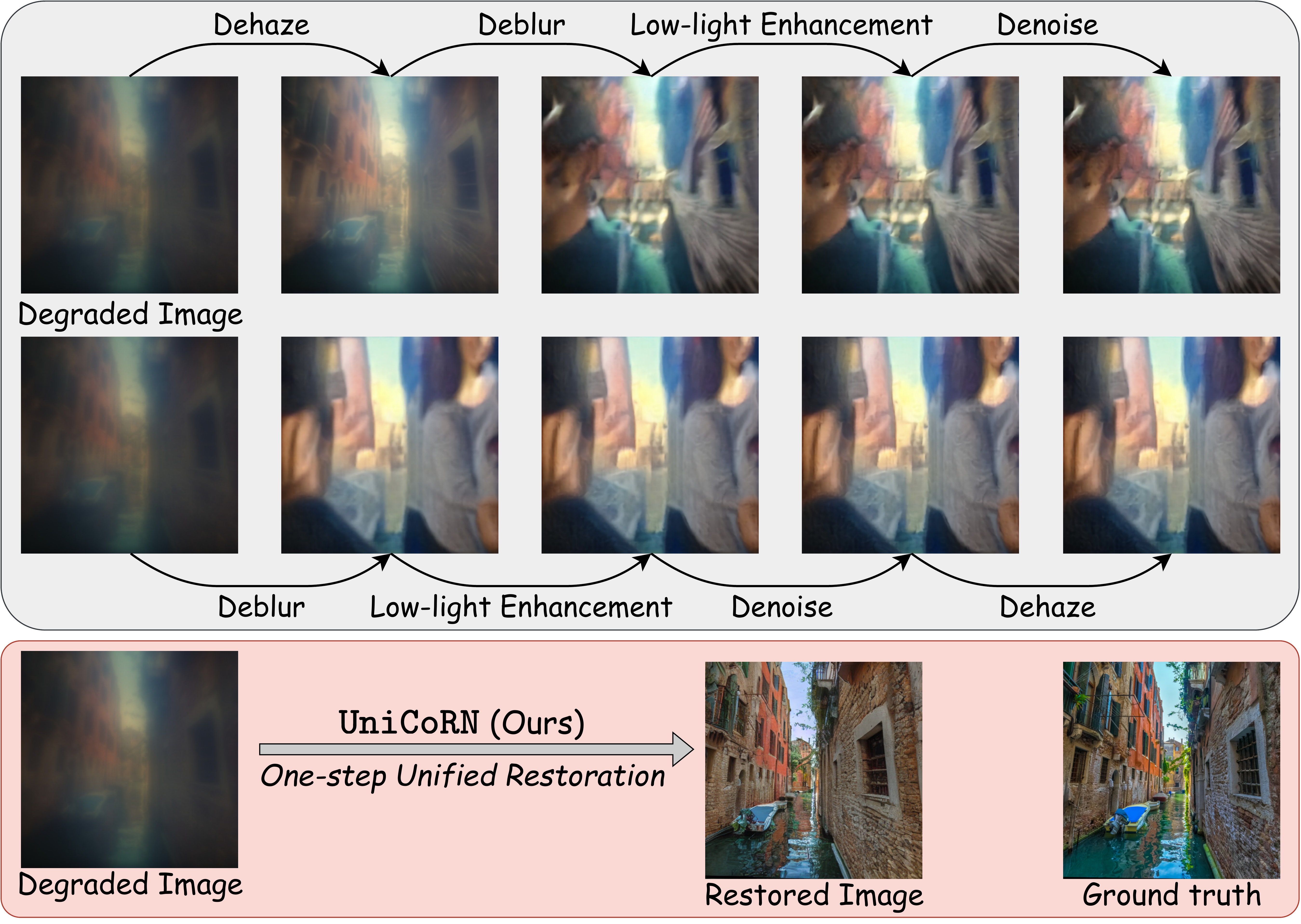}
    \caption{Applying a sequential combination of single degradation image restoration models still struggle to faithfully recover an image corrupted by multiple degradations. \modelName, on the other hand, can recover from multiple degradations, without prior knowledge of the corruption type.}
    \label{fig:pilot}
    \vspace{-0.5cm}
\end{figure}

In real-world scenarios, imaging systems often capture images in unconstrained, dynamic environments where multiple degradations may occur simultaneously, such as
autonomous vehicles facing multiple weather-related degradations \cite{wang2023effect}, underwater images affected by 
light scattering and absorption \cite{zhang2024underater, khan2024spectroformer}, or images from low-quality cameras \cite{wang2023effect, gao2021image,heide2013high} or flat optics \cite{khorasaninejad2017metalenses, tseng2018metalenses,tseng2021neural,chakravarthula2023thin} prone to multiple corruptions .
Unfortunately, the existing approaches \cite{zamir2022restormer, potlapalli2024promptir, li2020all} are mostly tailored 
for removing \textit{sole presence} of individual 
degradation types and \textit{do not generalize} well to handle multiple real-world corruptions simultaneously.
We ran a simple experiment of applying degradation-specific models \cite{zamir2022restormer, potlapalli2024promptir} sequentially
to remove simultaneous presence of multiple artifacts.
As shown in \cref{fig:pilot}, both approaches struggle to accurately restore the original image, often compounding artifacts and losing finer details.
This highlights the limitations of single degradation removal methods, and necessitates simultaneous
multi-degradation removal image restoration approaches.

To address the gap in multi-degradation image restoration, we propose \modelName, a unified model that simultaneously handles various types of degradations through robust artifact removal and high-fidelity image preservation.
Leveraging controllable diffusion \cite{rombach2022high} models as powerful generative priors \cite{ho2020denoising}, we model image restoration as a conditional generation task. 
Inspired by ControlNet \cite{controlnet}, we use a multi-head control network where each head targets a specific degradation type commonly seen in real-world images \cite{zhai2023comprehensive}.
The proposed \modelName~architecture allows us to utilize low-level\footnote{Low level vision refers to using image features such as edges, contours, color maps, transmission maps and depth maps} visual features \cite{chen2021edbgan, he2010single, yi2021structure, wang2022low} as degradation cues that condition image restoration and generation, providing better control than simple textual prompts \cite{controlnet, mou2024t2i}. 
To be precise, a
non-exhaustive set of low-level visual ``hints'', such as transmission, color or edge maps \cite{he2010single, joshi2009image, park2017low}, form the conditions for our control network, while a fixed generative diffusion model \cite{rombach2022high} serves as a backbone.
Furthermore, a mixture-of-experts \cite{shazeer2017outrageously, cai2024survey} strategy adaptively weights degradation cues, and 
curriculum learning \cite{bengio2009curriculum} enables the model to progressively learn handling both
single and compound degradations, especially mitigating the risk of catastrophic forgetting \cite{chen2018continual, kemker2018measuring} across tasks.

We have evaluated \modelName~on diverse single and multi-degradation image restoration datasets \cite{liu2021benchmarking, gopro, reside, cbsd68, div2k}, and demonstrate significantly improved performance over prior state-of-the-art methods \cite{zamir2022restormer, jiang2023autodir, potlapalli2024promptir, li2022all, zheng2024selective}.
Additionally, we introduce \dataName, a new benchmark dataset with metalens-captured images \cite{tseng2018metalenses} containing multiple real-world degradations and we further validate our model's robustness to real-world in-the-wild unconstrained photography challenges.

\noindent \textbf{Our core contributions, in summary, are as follows:}

\begin{enumerate}

    \item We introduce {\modelName}, a controllable latent diffusion model for unified image restoration that effectively recovers images with multiple, unknown degradations. Unlike most previous methods, our approach does not require prior knowledge of the specific degradations present.
    
    \item We design a multi-head control network that utilizes diverse low-level visual cues, adapting these through a mixture-of-experts strategy. We train the network with a curriculum learning approach to prevent catastrophic forgetting across various degradation removal tasks.

    \item We propose a new benchmark, \dataName, consisting of clean and metalens-captured degraded image pairs, to advance multi-degradation image restoration research.

    \item We demonstrate that our approach not only outperforms prior methods but also generalizes robustly to our new metalens imaging benchmark, achieving high efficacy in addressing diverse and challenging artifacts.
    
\end{enumerate}

\noindent Code and data will be made public upon acceptance.

\def\red#1{\textcolor[rgb]{1,0,0}{#1}}
\def\blue#1{\textcolor[rgb]{0,0,1}{#1}}
\def\orange#1{\textcolor[rgb]{1,0.49,0}{#1}}
\def\yellow#1{\textcolor[rgv]{1,1,0}{#1}}
\def\etal{\emph{et al}.}
\def\modelName{\textsc{UniCoRN}}
\def\dataName{\textsc{MetaRestore}}
\section{Related Works}

\label{sec:related}

\keypoint{Image Restoration for Single Degradations:}A bulk of image restoration approaches have focused on \textit{single degradation removal}, i.e. when an image is assumed to have only one form of (known) degradation. These include denoising \cite{buades2005non, lehtinen2018noise2noise, fan2019brief, joshi2009image}, dehazing \cite{song2023dehazeformer, dong2024dehazedct, li2021you, lee2016review, li2020zero, xu2012fast}, deblurring \cite{zhang2022deep, cho2021rethinking, tao2018scale, yan2017image, joshi2009image}, low-light enhancement \cite{liu2021benchmarking, park2017low, ren2016image, yang2023lightingnet, malik2023semi}, among others. Traditional methods have relied on color \cite{joshi2009image}, dark channel \cite{xu2012fast, he2010single}, bright channel \cite{park2017low} and low-rank \cite{ren2016image} priors, which are all specific to a given degradation type and hence they cannot be applied across different degradations. More recently, deep learning models for image restoration \cite{song2023dehazeformer, zamir2022restormer, dong2024dehazedct, li2020zero, wang2022uformer, liang2021swinir} have emerged as superior alternatives by enabling end-to-end mapping of degraded input images to their clean counterparts, and being agnostic to the degradation removal task. Such methods include Restormer \cite{zamir2022restormer}, pyramid attention networks \cite{mei2023pyramid}, SwinIR \cite{liang2021swinir}, U-Former \cite{wang2022uformer} and DoubleDIP \cite{gandelsman2019double}, to name a few. These methods offer a common architecture for image restoration; however, they have to be trained separately for each type of degradation, requiring separate instances for tackling distinct degradations \cite{zamir2022restormer, liang2021swinir}. This makes the above approaches computationally complex \cite{valanarasu2022transweather} and unsuitable for practical usage where multiple (and often unknown) degradations co-exist in images. In contrast, our work not only generalizes across various degradations but removes them without any prior assumption of specific degradations.

\keypoint{Image Restoration for Multiple Degradations:} Departing from single-degradation image restoration, there have been attempts to model multiple degradations jointly \cite{valanarasu2022transweather, li2020all, li2022all, chen2021pre, luo2023controlling} by \textit{learning degradation-aware modules} \cite{chen2021pre, potlapalli2024promptir, li2022all, valanarasu2022transweather}. These include Li \etal \cite{li2020all} which employs neural architecture search, Pre-trained Image Processing Transformer (IPT), \cite{chen2021pre} employing multiple heads and tails for different degradations with a shared backbone, TransWeather \cite{valanarasu2022transweather} which learns dynamic weather embeddings, and AirNet \cite{li2022all}, employing contrastive learning for degradation-aware discriminative features. Another set of works leverage prompt learning \cite{nlp_prompting, chattopadhyay2024towards} to dynamically encode degradation-aware context \cite{potlapalli2024promptir, li2023prompt, ma2023prores}. However, most of these approaches have been limited to weather degradations only \cite{li2020all, valanarasu2022transweather, rajagopalan2024awracle, potlapalli2024promptir} and typically cannot tackle multiple degradations simultaneously. More recently, a few works have attempted universal image restoration \cite{li2022all, jiang2023autodir, zheng2024selective}, modeling degradations without any prior assumptions. 
Our work proposes such a generalized image restoration model to \textit{remove co-existing multiple degradations in a single step}, a direction heavily under-explored in literature.

\keypoint{Conditional Diffusion Models:} Diffusion models \cite{ho2020denoising, kawar2022denoising, rombach2022high, controlnet} are becoming the de facto controllable visual content generation frameworks for various tasks that include images \cite{rombach2022high}, videos \cite{blattmann2023align} and 3D objects \cite{poole2022dreamfusion}. Popular diffusion models include Stable Diffusion \cite{rombach2022high}, T2I-Adapter \cite{mou2024t2i}, ControlNet \cite{controlnet} and DreamBooth \cite{ruiz2023dreambooth}, among others. 
Furthermore, conditional diffusion models allow precise control over the denoising process to generate outputs customized to the user's needs, be it through textual prompts \cite{rombach2022high}, low-level guidance such as edge/depth/color maps \cite{controlnet, zhao2024unicontrolnet}, or via personalization \cite{ruiz2023dreambooth}. 
In particular, image restoration has also been attempted using controllable diffusion models \cite{he2024diffusion, zheng2024selective, luo2023controlling}, where the \textit{conditioning signal typically contains the degradation-relevant information.} DA-CLIP \cite{luo2023controlling} and D4IR \cite{wang2024data} train a degradation-aware encoder via contrastive learning \cite{clip}, the output of which is then used to prompt the diffusion model. DiffUIR \cite{zheng2024selective} aims at bridging distributions of different degradation tasks, while AutoDIR \cite{jiang2023autodir} identifies the dominant degradation in an image and conditions a structural-corrected diffusion model for restoration. 
Our work goes beyond vanilla spatial or semantic controllability of diffusion models and instead it proposes the use of stronger priors for degradation removal in the form of a \textit{non-exhaustive set of low-level visual cues} that can be easily extracted from a given (degraded) image.

\def\modelName{\textsc{UniCoRN}}
\def\dataName{\textsc{MetaRestore}}

\def\red#1{\textcolor[rgb]{1,0,0}{#1}}
\def\blue#1{\textcolor[rgb]{0,0,1}{#1}}
\def\black#1{\textcolor[rgb]{0,0,0}{#1}}
\def\orange#1{\textcolor[rgb]{1,0.49,0}{#1}}
\def\yellow#1{\textcolor[rgv]{1,1,0}{#1}}
\def\etal{\emph{et al}.}
\def\modelName{\textsc{UniCoRN}}
\def\dataName{\textsc{MetaRestore}}

\section{Methodology}

Our method uses low-level visual cues to restore images corrupted by multiple degradations, simultaneously.
Here, we begin with a discussion on multi-degradation image formation and extracting visual guidance cues.
We then continue with our unified image restoration network and its task-hierarchical curriculum learning training strategy.

\begin{figure*}[htbp]
    \centering
    \includegraphics[width=\textwidth]{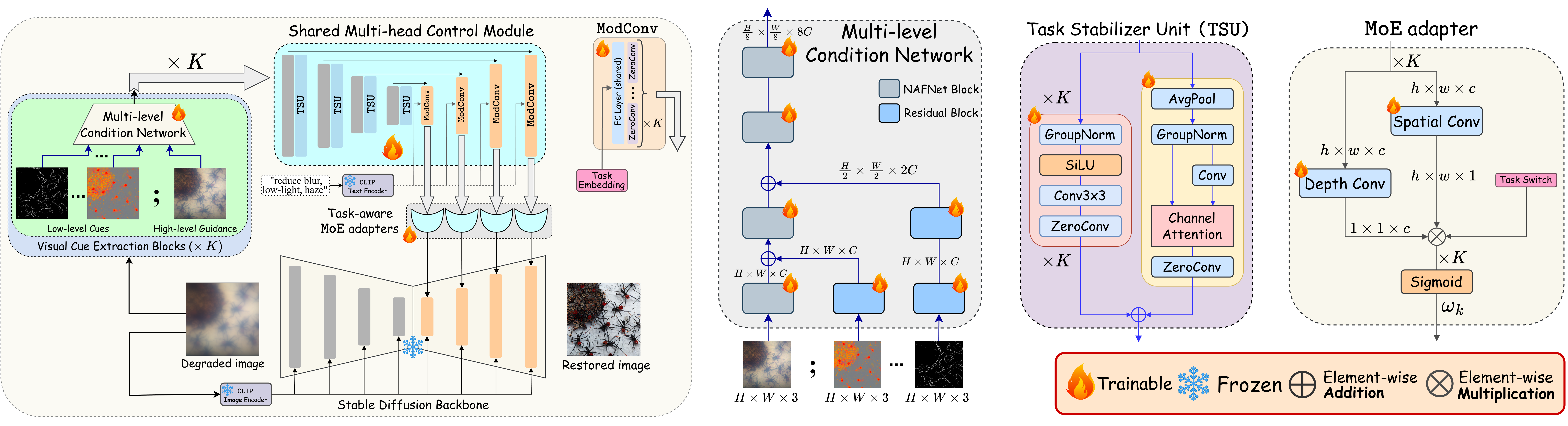} %
    \caption{Overview of our proposed {\modelName} model for unified multi-degradation image restoration.}
    \label{fig:overview} %
\end{figure*}

\subsection{Multi-degradation Image Formation}
For a given source image $S$, the imaging process for a camera with a wavelength-dependent point spread function (PSF) $p_\lambda$ can be described as:

\vspace{-5mm}
\begin{equation}
    I_c(x, y) = \mathcal{P}\Big( \int (\rho_\lambda(S) * p_\lambda)(x,y) \kappa_c(\lambda) \ d\lambda \Big) + \mathcal{N}
    \label{eq:unknown_image_formation}
\end{equation}
where $I_c\in\mathbb{R}^{H\times W \times C}$ represents the sensor measured image with $c$ denoting the RGB channels, $\rho_\lambda(\cdot) < 1$ is the lens's relative light efficiency, $\kappa_c(\lambda)$ the spectral sensitivity of channels to wavelength $\lambda$, and $*$ denotes convolution.
$\mathcal{P}(\cdot)$ and $\mathcal{N}$ represents Poisson noise and additive Gaussian noise, respectively. 
The source image ($S$) can then be estimated as  
$\hat{S} = \arg\max_{S} \probP(S|I) = \arg\max_S \probP(I|S)\probP(S)$ 
using a \textit{maximum a-posteriori} framework with a known prior $\probP(S)$.

\vspace{2mm}
\noindent
\textbf{Restoration} of image $S$ from sensor measurement $I$ introduces severe artifacts due to the ill-posed nature of the image formation process (\cref{eq:unknown_image_formation}), especially impacted by the sensor's spectral response $\kappa_c(\lambda)$ and noise $\mathcal{N}$.
Ignoring $\kappa_c(\lambda)$ reduces the image restoration problem to deblurring with a known PSF, but integrating wavelength-dependent $\kappa_c$ over $\lambda$ introduces haze in the image, requiring both deblurring and dehazing.
Sensor noise further degrades high frequencies in the image, leading to loss of fine details.
Therefore, to simplify the restoration problem, we pose \cref{eq:unknown_image_formation} as

\vspace{-1.5mm}
\begin{equation}
    I(x, y) = \mathcal{H}(\mathcal{B}(\mathcal{D}(S))) + \eta
    \label{eq:compact}
\end{equation}
where $\mathcal{H}(\cdot)$, $\mathcal{B}(\cdot)$, $\mathcal{D}(\cdot)$ and $\eta$ represent hazing, blurring, image darkening and the noise.  
Note that reconstructing $S$ from simplified \cref{eq:compact} still remains a challenging non-linear problem, beyond the scope of traditional deblurring, denoising, dehazing or contrast enhancement techniques.

\subsection{Extracting Low-level Restoration Cues}
\label{sec:cues}

The initial step of our restoration pipeline to solve \cref{eq:compact} is to extract essential low-level visual cues from the degraded input image. 
Given an image $I$ 
and a task $h \in \{\mathcal{H}, \mathcal{B}, \mathcal{D}, \eta\}$, we generate structural guidance estimates \textbf{(secondary cues)} $I^h_{\{ 1,2,...\}}$ along with initial image estimates \textbf{(primary cues)} $I^h_0$ for each task $h$ (see \autoref{fig:overview}) using cheap and fast blind restoration methods. 

\keypoint{Secondary cues.}These involve the following:

\begin{itemize}
\item \textit{Dark channel prior-aided transmission map.} Using the dark channel prior (DP) of the image \cite{he2010single}, we obtain:

\vspace{-1.5mm}
\begin{equation}
I_1^{\mathcal{H}} = t(x) = 1 - \omega\cdot \,\mbox{DP}(I/A).
\end{equation}

\item \textit{Edge map.} An edge map estimation ($I_1^{\mathcal{B}} = e$)
constructed from the shock filtered \cite{cho2009fast} image ($I_2^{\mathcal{B}}$) with edge thresholding:

\vspace{-1.5mm}
\begin{equation}
    e_{t+1} = e_t - \texttt{sign}(\Delta e_t)\|\nabla e_t\|dt
\end{equation}

where $e_t$ is the edge map at time t ($e_0 \!= \!I$), $\Delta e_t$ and $\nabla e_t$ are the Laplacian and gradient respectively, and $dt$ is the time step for a single filtering evolution. Additionally, following Perona-Malik anisotropic diffusion \cite{perona1994anisotropic}, we extract a \textit{noisy edge map} ($I_1^{\mathcal{\eta}}$) from running diffusion at two different conduction coefficients for additional guidance.

\item \textit{Color map.} We compute a basic color map ($I_1^{\mathcal{D}} = m$) as secondary guidance inspired by Retinex Theory:

\vspace{-1.5mm}
\begin{equation}
   m(I) = \frac{3 \cdot I}{\sum_{c}I_c}
\end{equation}

\end{itemize}

\keypoint{Primary cues.} The initial image estimates come from the following:

\begin{itemize}

\item \textit{Haze-removal estimate.} We can obtain an initial haze-free image from the relation $I(x) = I(x)_0^{\mathcal{H}} \cdot t(x) + A \cdot (1 - t(x))$, where $A$ is the estimated atmospheric light level. To optimize performance, we replaced the soft matting operation from \cite{he2010single} with a guided filtering method \cite{he2012guided}.

\item \textit{Blur removal estimate.} We use Wiener deconvolution for the primary guidance $I_0^{\mathcal{B}}$ in the presence of a known PSF blur kernel.

\item \textit{Contrast-enhanced estimate.} We apply adaptive histogram equalization \cite{pizer1987adaptive} for contrast enhancement to generate an initial contrast-enhanced estimate $I_0^{\mathcal{D}}$.

\item \textit{Smoothened image estimate.} We use Perona-Malik anisotropic diffusion \cite{perona1994anisotropic} with reference to \citet{gberg}, to estimate ($I_0^{\mathcal{\eta}}$) edge-preserving smoothed image with significantly faster computation times.
    
\end{itemize}

\subsection{Unified Controllable Restoration Network} 
\label{sec:unicorn}
In this section, we introduce a novel conditioning network inspired by the ControlNet \cite{controlnet} architecture based on Latent diffusion models.
First, we present a powerful multi-level condition network (MLCN) to encode and aggregate guidance across multiple stages, which is then injected into the locked diffusion model's layer outputs.
Next, we describe our shared multi-headed control module, augmented with a task-aware mixture-of-experts (MOE) adapter that effectively mixes control signals. Additionally, we incorporate a 
task-wise modulated layer at various \black{control} levels to further refine the conditioning. 
Finally, we explore a gradient stabilization mechanism for task-switching during training.

\keypoint{Multi-Level Condition Network (MLCN).} 
For a specific task $h$ and a set of input image guidances \{$I_{0, 1, 2, ...}^h$\}, we engineer a cascaded feature encoder structure with multi-level feature aggregation.
One way to combine multiple guidances would be to stack them together and pass through a single learnable layer.
However, this method has two key limitations: 
first, the primary guidance $I_0^h$  significantly differs in visual characteristics from the lower-level guidances $I_{1,2,...}^h$ and secondly, features of varying complexities must be learned for each set. 
Second, some guidances can be learned more effectively using a \textit{shallower encoder} than a deeper one.
Capitalizing on these insights, we designed our multi-level condition encoder, as illustrated in \cref{fig:overview}.

Our MLCN consists of a main encoder branch $\phi_0$ for primary guidance and secondary branches $\phi_{1,2...}$ for low-level guidances. At the initial level i.e. $j=0$, for every task $h$ we have $x^0_{0,1,...} = I^h_{0,1,...} \in \mathbb{R}^{H\times W\times 3}$. At each encoder level j, the features $x^{j-1}$ from the previous layers are downsampled by a factor of $2$ and aggregated with those from the secondary guidances at each level. This process can be denoted as:

\vspace{-2mm}
\begin{equation}
    x_b^j = \phi_b^j(x_b^{j-1}) + \mathbbm{1}_{\{ b = 0 \}} \cdot \sum\limits_{r \in \Omega_j} \phi_r^j(x_r^{j-1})
\label{eq:optprob}
\end{equation}
for each $b$ in the full set of guidances $\{0,1,...\}$. $\Omega_j$ contains hints $r$ for which the final level equals $\mathbf{j}$ indicating which hints need to be aggregated at that particular level. At the final level $(L=4)$, a single input control tensor $C^0_h$ is obtained from the primary guidance branch $\phi^L_0$ for each task $h$. For the secondary guidances $I^h_{1,2...}$, we choose residual encoder blocks for $\phi$ due to its shortcut paths providing a shallower depth. For the primary task guidances $I^h_0$ we choose the computationally efficient NAFBlock \cite{chen2022simple} for degradation invariant feature representation.

\keypoint{Shared Multi-Head Control Module.}The advantage of learning separate heads instead of one across tasks, unlike \cite{qin2023unicontrol}, is its potential to capture feature representations separately from one another. While this can be viewed as a bottleneck to expanding to broader degradation sets, it is mitigated by the efficiency of using a single shared Stable Diffusion \cite{rombach2022high} control head across all tasks (see \cref{fig:overview}). 
The controls $C^j_{1...K}$ themselves go through a task-specific convolutional block and a \textit{task stabilization unit} (TSU) (described below) before passing through the next control layer, 

\vspace{-5mm}
\begin{align}
c^{j}_{h} &= C^j_h + \texttt{TSU}(\{C^j_{1...K}\}) + \texttt{TaskBlock}(C^j_h) \label{eq:gsusum}\\
C^{j+1}_h &= \mathcal{E}^j_{control}(c^j_{h}, e^{img\downarrow_4}_{clip}, t),\label{eq:mu}
\end{align}

where, $\mathcal{E}^j_{control}$ denotes the shared control encoder at level $j$, $e^{img\downarrow_4}_{clip}$ is the downsized LQ image CLIP \cite{clip} embedding for semantic guidance and $t$ denotes the diffusion timestep \cite{ho2020denoising}. 

Given the control information $C^j_h$ from each task $h$ at a control level $j$ and we compute the overall control injection into the frozen diffusion model using a lightweight mixture-of-experts (MOE) \cite{shazeer2017outrageously} adapter,
\begin{equation}
    {C}^j_{moe} = \texttt{MOE-Adapter}(\{C^j_{1...K}\})
\end{equation}

\noindent
Using ${C}^j_{moe}$ as the final control signal for the controllable LDM, we can write the loss objective as

\vspace{-2.5mm}
\begin{equation}
    \resizebox{\columnwidth}{!}{
    $\mathcal{L} = \mathbb{E}_{\mathbf{z}_0, \mathbf{t}, \mathbf{e}^{img\downarrow_4}_{clip}, \mathbf{C}^j_{moe}, \epsilon \sim \mathcal{N}(0,1)} [ \lVert \epsilon - \epsilon_{(\phi, \mathcal{E})}(\mathbf{z}_t, \mathbf{t}, \mathbf{e}^{img\downarrow_4}_{clip}, \mathbf{C}^j_{moe}) \rVert^2_2]$
    }
\end{equation}

\keypoint{Task Stabilizer Unit for switching tasks.} 
When the model transitions training over the set of possible single and mixed tasks many times within an epoch, the control encoder layers ($\mathcal{E}$) are forced to juggle between learning different tasks. This destabilizes the training process and to avoid this, we introduce a task stabilization layer in the form of a shared block between every pair of control encoder blocks as shown in \cref{fig:overview}. Rewriting the TSU term from \cref{eq:gsusum},
\begin{equation}
    \texttt{TSU}(\{C^j_{1...K}\}) = \texttt{ZC}\uparrow(\texttt{CAT}(\texttt{Conv}\downarrow(\texttt{GN}(C^j_{avg}))))
    \label{eq:stablegrad}\notag
\end{equation}
where, $C^j_{avg} = \frac{1}{|H|}\sum_hC^j_h$ and \texttt{GN}, \texttt{Conv}, \texttt{CAT} and \texttt{ZC} stand for group normalization, convolution, channel attention and zero convolution \cite{controlnet} layer respectively. The TSU block's input is the average over all task controls which reduces to just one control $C^j_h$ in the case of single tasks. During task switching the TSU block helps modulates the current gradient from the previously learned tasks due to its residual form.

\keypoint{Task-specific control blocks} 
Further, we provide additional task dependent \textit{memory} in between the encoder layers as a task specific depthwise separable convolutional layer. 

\vspace{-4mm}
\begin{equation}
    \texttt{TaskBlock}(C^j_h) = \texttt{ZC}_{point} (\texttt{Conv}_{depth}(\texttt{SiLU}(\texttt{GN}(C^j_h))))\notag
\end{equation}
Here, $\texttt{ZC}_{point}$  and $\texttt{Conv}_{depth}$ denote pointwise ($1\times1$ kernel) and depthwise convolutions respectively, with SiLU activation \cite{silu} layer.

\paragraph{Task-aware MOE adapter.} 
We model a mixture block for weighting the control paths $C^j_h$ based on a separable convolution layer, placed at the beginning on each control injection ($C^j_{moe}$) into the diffusion model (\cref{fig:overview}),
\begin{equation}
    C^j_{moe} = W(\{C^j_h\}, p^h_{clip}) = \sum\limits_h S_hw^j_h(C^j_h, p^j_{clip}) C^j_h
\end{equation}
where, $p^h_{clip}$ is the textual task prompt CLIP embedding for the particular degradation task and $S_h \in \{0,1\}$ is a user-provided task switch.

To keep the overall control head lightweight, we turn the weighting layer (W) into a separable convolutional layer as shown in \cref{fig:overview} having the following form,
\begin{equation}
    w^j_h(M^j_h) = \texttt{SConv}(M^j_h).\texttt{DConv}(M^j_h)
\end{equation}
where $M^j_h = \texttt{ModConv}(C^j_h, p^j_{clip})$ and, $\texttt{SConv}$ and $\texttt{DConv}$ provide independent spatial $(h,w,c)\!\to\!(1,1,c)$ and channel wise $(h,w, c)\!\to\!(h,w,1)$ averaging respectively. Here, $\texttt{ModConv}$ is a task based modulation convolution inspired from the StyleGAN2 literature \cite{karras2020analyzing}.

\begin{figure*}[!htbp]
	\captionsetup{font=small}
	\scriptsize
	\centering
	
	\newcommand{\h}{0.105}
	\newcommand{\wa}{0.12}
	\newcommand{\wb}{0.16}
	\newcommand{\g}{-0.7mm}

 	\setlength\tabcolsep{1.8pt}
	\renewcommand{\arraystretch}{1}
	\resizebox{1.00\linewidth}{!} {
			\renewcommand{\h}{0.15}
			\newcommand{\w}{0.200}
				\begin{adjustbox}{valign=t}
					\begin{tabular}{cccccccc}
						\includegraphics[height=\h \textwidth, width=\w \textwidth]{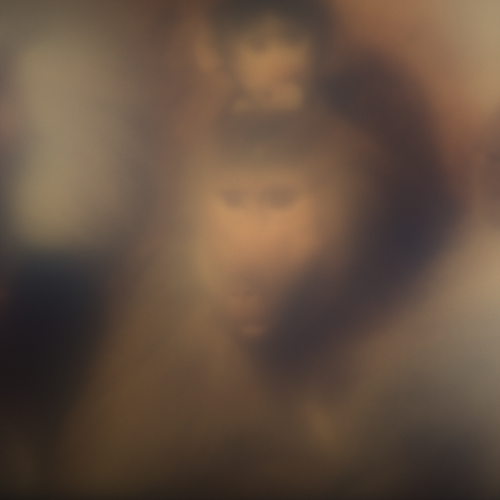} \hspace{\g} &
						\includegraphics[height=\h \textwidth, width=\w \textwidth]{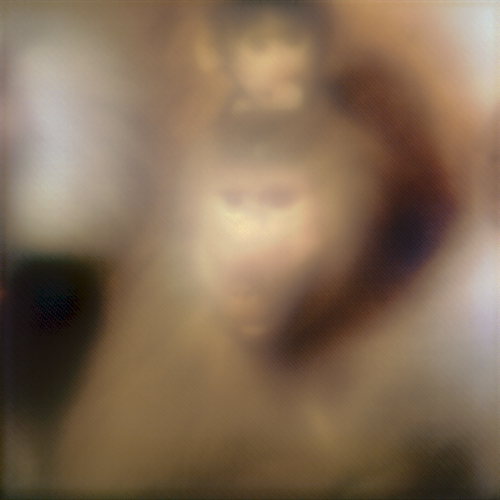} \hspace{\g} &
                        \includegraphics[height=\h \textwidth, width=\w \textwidth]{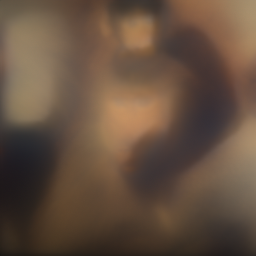} \hspace{\g} &
                        \includegraphics[height=\h \textwidth, width=\w \textwidth]{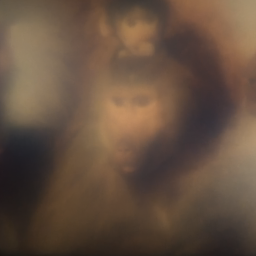} \hspace{\g} &
                        \includegraphics[height=\h \textwidth, width=\w \textwidth]{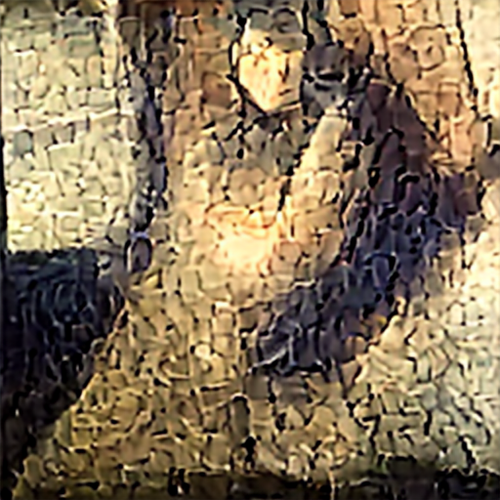}
                        \hspace{\g} &
                        \includegraphics[height=\h \textwidth, width=\w \textwidth]{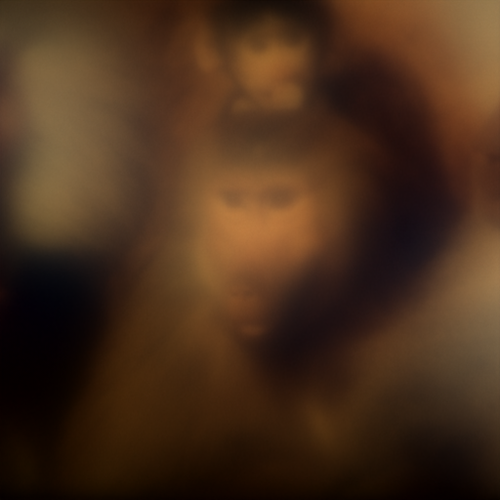}
                        \hspace{\g} &
                        \includegraphics[height=\h \textwidth, width=\w \textwidth]{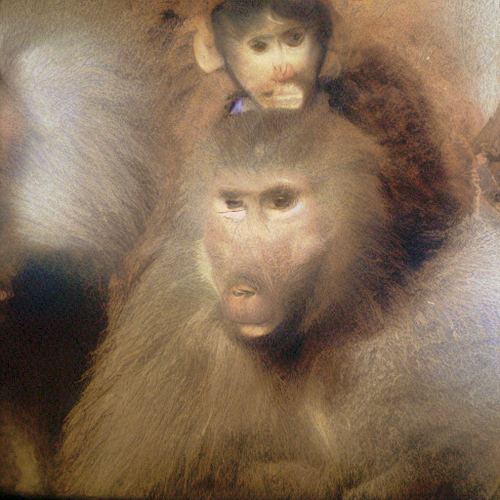} \hspace{\g} &
                        \includegraphics[height=\h \textwidth, width=\w \textwidth]{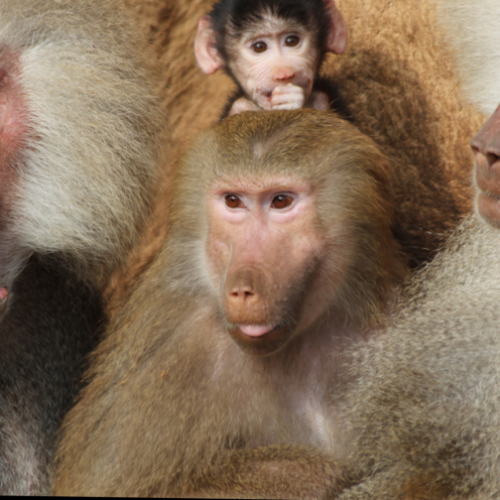}
                        \\
                        \includegraphics[height=\h \textwidth, width=\w \textwidth]{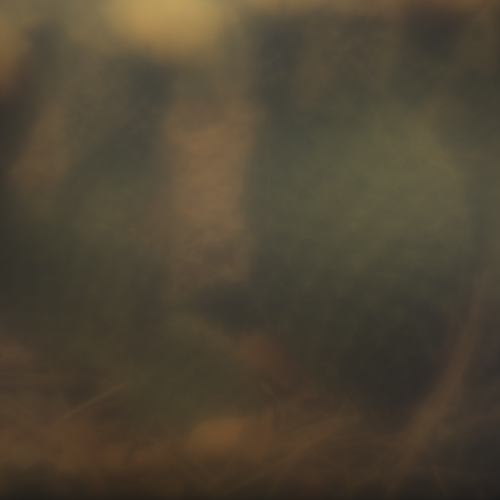} \hspace{\g} &
						\includegraphics[height=\h \textwidth, width=\w \textwidth]{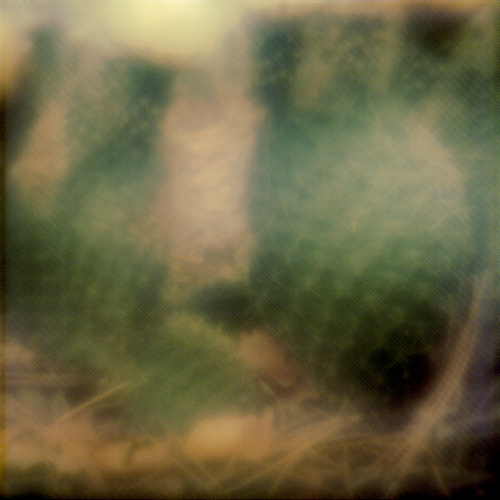} \hspace{\g} &
                        \includegraphics[height=\h \textwidth, width=\w \textwidth]{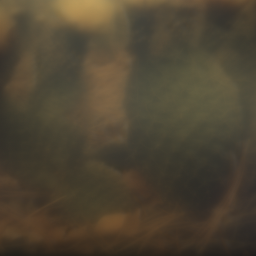} \hspace{\g} &
                        \includegraphics[height=\h \textwidth, width=\w \textwidth]{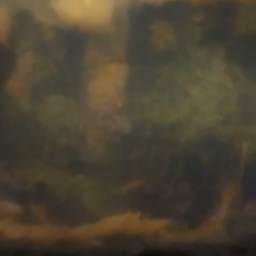} \hspace{\g} &
                        \includegraphics[height=\h \textwidth, width=\w \textwidth]{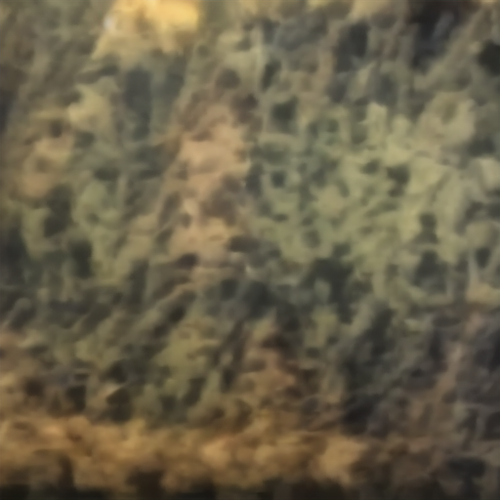}
                        \hspace{\g} &
                        \includegraphics[height=\h \textwidth, width=\w \textwidth]{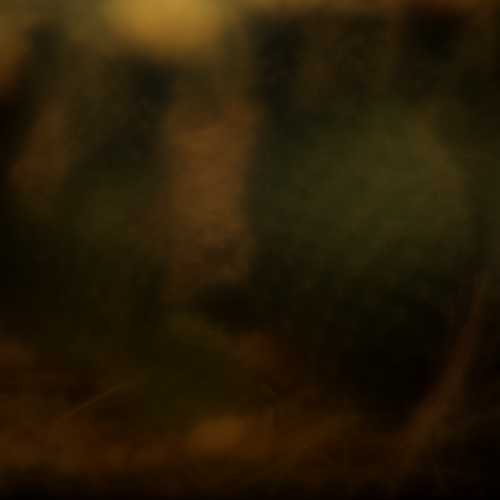}
                        \hspace{\g} &
                        \includegraphics[height=\h \textwidth, width=\w \textwidth]{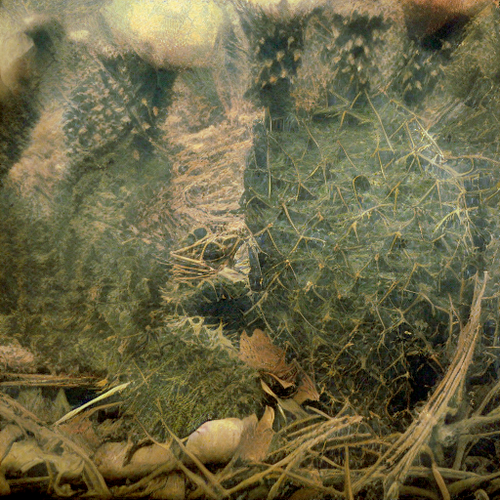} \hspace{\g} &
                        \includegraphics[height=\h \textwidth, width=\w \textwidth]{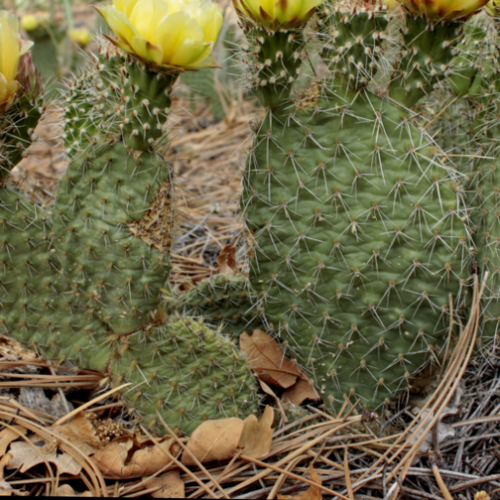}
                        						\\
                        \includegraphics[height=\h \textwidth, width=\w \textwidth]{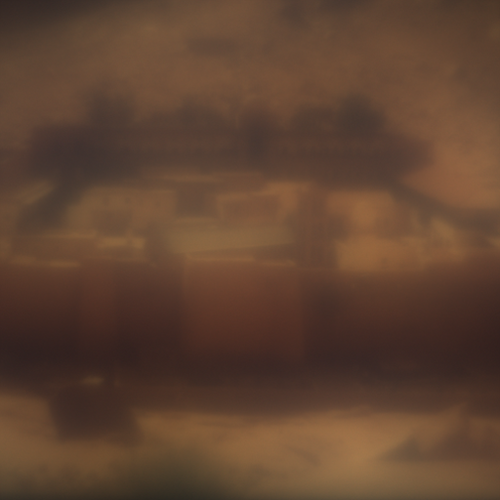} \hspace{\g} &
						\includegraphics[height=\h \textwidth, width=\w \textwidth]{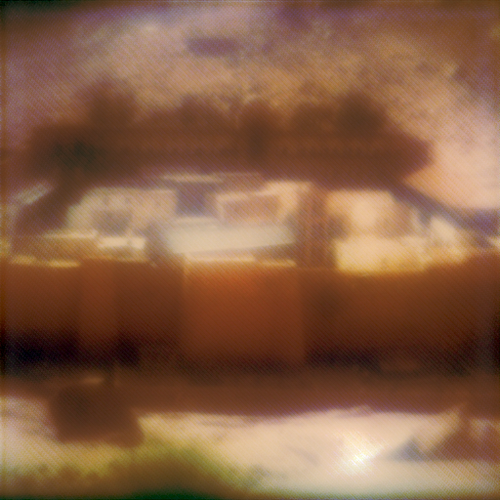} \hspace{\g} &
                        \includegraphics[height=\h \textwidth, width=\w \textwidth]{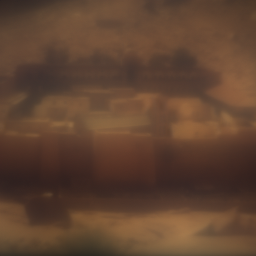} \hspace{\g} &
                        \includegraphics[height=\h \textwidth, width=\w \textwidth]{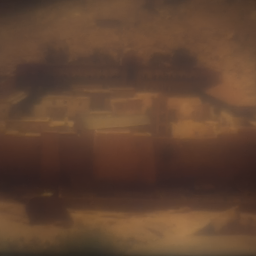} \hspace{\g} &
                        \includegraphics[height=\h \textwidth, width=\w \textwidth]{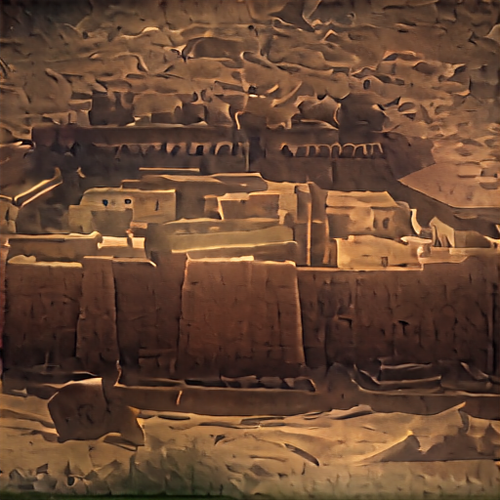} \hspace{\g} &
                        \includegraphics[height=\h \textwidth, width=\w \textwidth]{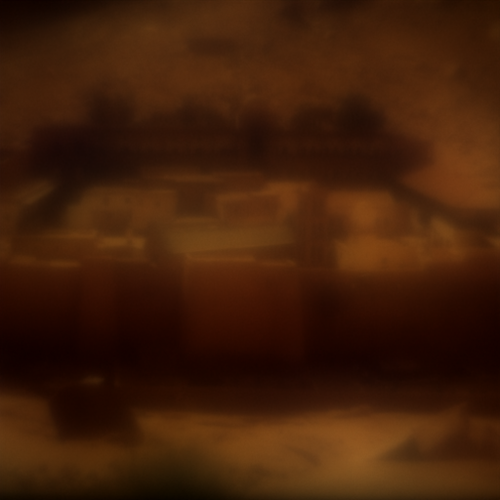}
                         \hspace{\g} &

                        \includegraphics[height=\h \textwidth, width=\w \textwidth]{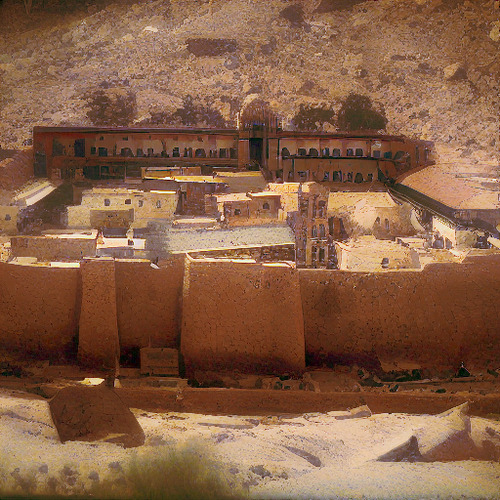} \hspace{\g} &
                        \includegraphics[height=\h \textwidth, width=\w \textwidth]{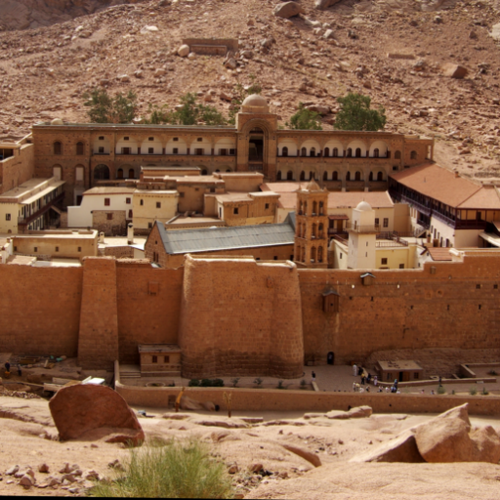}
                        \\
						\includegraphics[height=\h \textwidth, width=\w \textwidth]{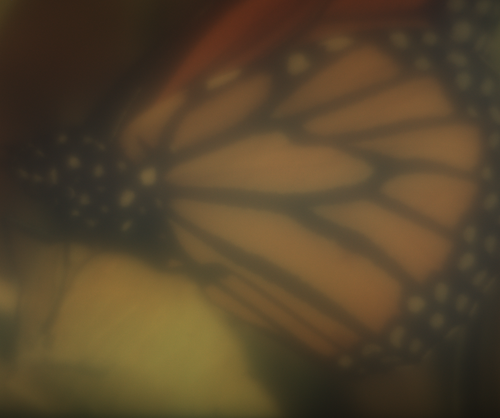} \hspace{\g} &
						\includegraphics[height=\h \textwidth, width=\w \textwidth]{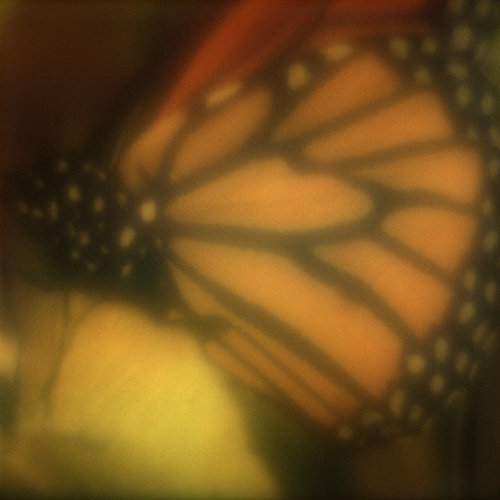} \hspace{\g} &
                        \includegraphics[height=\h \textwidth, width=\w \textwidth]{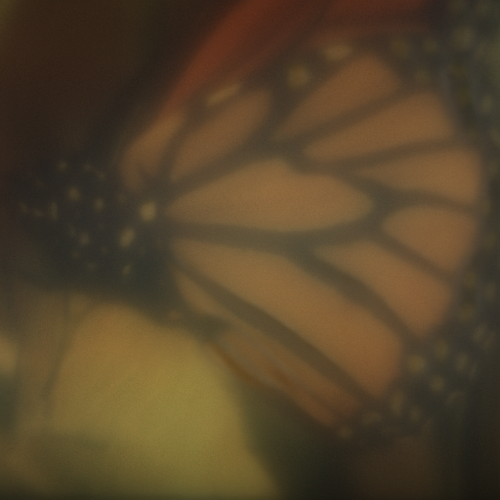} \hspace{\g} &
                        \includegraphics[height=\h \textwidth, width=\w \textwidth]{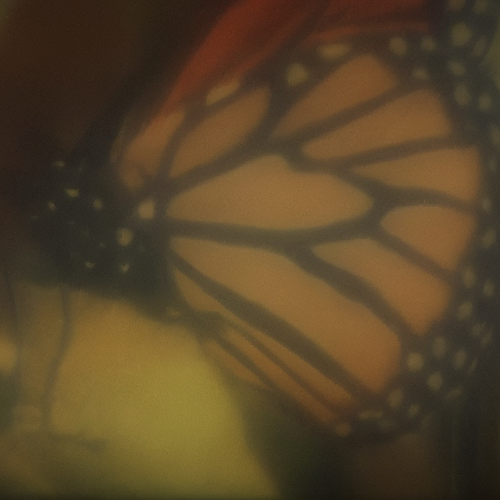} \hspace{\g} &
                        \includegraphics[height=\h \textwidth, width=\w \textwidth]{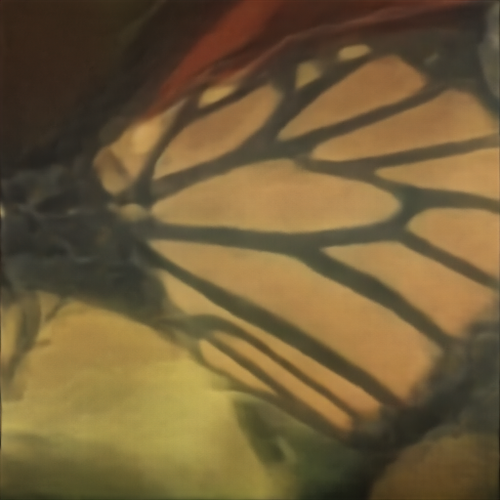} \hspace{\g} &
                        \includegraphics[height=\h \textwidth, width=\w \textwidth]{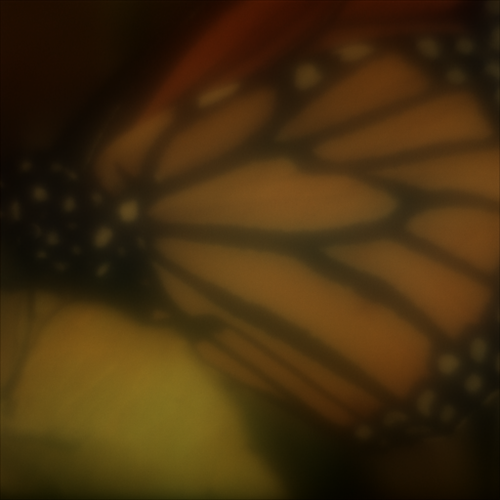} \hspace{\g} &
                        \includegraphics[height=\h \textwidth, width=\w \textwidth]{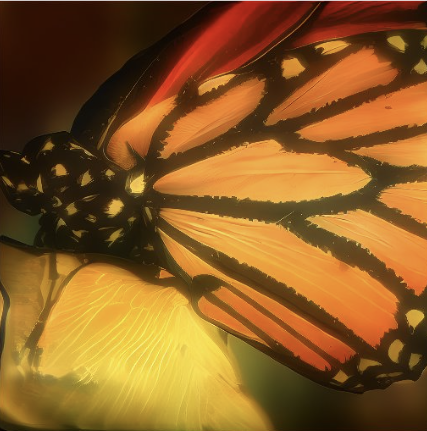} \hspace{\g} &
                        \includegraphics[height=\h \textwidth, width=\w \textwidth]{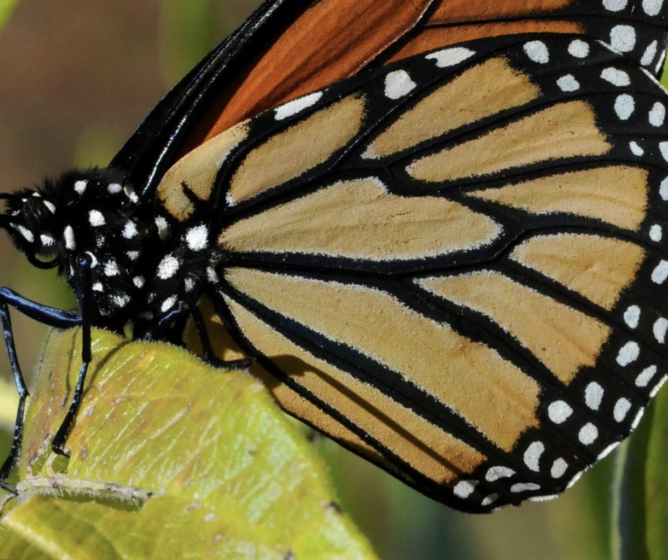}
						\\
						\textbf{Metalens Capture} \hspace{\g} &
                        \textbf{AirNet}~\cite{li2022all}  \hspace{\g} & 
                        \textbf{DiffUIR-L}~\cite{zheng2024selective}  \hspace{\g} & 
                        \textbf{DA-CLIP}~\cite{luo2023controlling}  \hspace{\g} & 
                        \textbf{AutoDIR}~\cite{jiang2023autodir}  \hspace{\g} & 
                        \textbf{PromptIR}~\cite{potlapalli2024promptir}  \hspace{\g} & 
                        \textbf{Ours}  \hspace{\g} & 
                        \textbf{GT}

					\end{tabular}
				\end{adjustbox}
		 }

 \vspace{-3mm}
	\caption{Qualitative results on the proposed {\dataName} metalens imaging benchmark. The results shown here are \textbf{\textit{zero-shot}}; none of the models have been fine-tuned on the dataset. Best viewed if zoomed in.}
	\label{fig:metalens_zeroshot}
 \vspace{-3mm}
\end{figure*}

\subsection{Task-hierarchical Curriculum Learning}
\label{sec:taskl}
We employ a curriculum learning \cite{bengio2009curriculum} process to augment the training of the control layers. In order to preserve the diversity of each control path we train on single degradation tasks $h$ and then progressively train on a few multi-degradation tasks to improve the generalization capabilities of individual heads. This separation between tasks helps us categorize them based on difficulty levels and train the model with an incremental learning \cite{chen2018continual} curve. Mixed degradation tasks, though derived from single degradations, better reflect real-world scenarios but challenge the model in early training. Hence, \textit{we topologically sort both single and synthetic multi-degradation datasets} based on a \textit{parent-child relationship graph} and, \textit{further sort datasets within a level in the descending order of size} to minimize the effect of \textit{catastrophic forgetting} \cite{chen2018continual, kemker2018measuring} \textit{of smaller task datasets}.

\def\modelName{\textsc{UniCoRN}}
\def\dataName{\textsc{MetaRestore}}

\section{Experiments}
\label{sec:experiments}

\keypoint{Implementation.}Leveraging frozen Stable Diffusion v1.4 \cite{rombach2022high}, we train {\modelName} using the curriculum learning recipe described in \cref{sec:taskl}, on a combined corpus of single and multi-degradation datasets (having paired combinations), with a learning rate of $1\mathrm{e}{-5}$, AdamW \cite{loshchilov2017decoupled} optimizer, accelerated by 2 NVIDIA A100 40GB GPUs, for approx. 120h (until convergence). We implement the model in PyTorch \cite{paszke2019pytorch} and use gradient accumulation \cite{lin2018deep} over $8$ batches to boost training efficiency. The CLIP encoders \cite{clip} have embedding dimension $d=768$ and remain frozen.

\keypoint{{\dataName} benchmark.} 
We position a metalens camera one meter from an LCD display, optimizing exposure, white balance, and contrast. In a dark room, the camera captures $800$ Div2K \cite{div2k} training images sequentially, controlled by a Linux server via Vimba SDK. Similarly, $400$ images were captured from \cite{Liu4K} for evaluation. \textit{This setup yields heavily degraded images due to metalens limitations} \cite{tseng2018metalenses} (see supplementary material for more details).

\keypoint{Additional Datasets.}We evaluate \modelName{} on benchmarks for \underline{mixed synthetic datasets}: Blur+Noise (Div2k \cite{div2k} with 2-30\% AWGN and random blur kernels),  LOL+Blur \cite{zhou2022lednet} and a synthetic Haze+Blur dataset constructed from haze images blurred using random blurring kernels. Furthermore, we also test on \underline{single-degradation removal} datasets: GoPro \cite{gopro} for motion deblurring, RESIDE \cite{reside} for outdoor dehazing, LOL \cite{wei2018deep} for low light enhancement, and CBSD68 \cite{cbsd68} for image denoising.

\keypoint{Evaluation Metrics.}We use \text{PSNR} \cite{psnr} and \text{SSIM} \cite{ssim} for distortion evaluation, and \text{LPIPS} \cite{lpips} for perceptual assessment. Additionally, we use non-reference image metrics \text{NIQE} \cite{niqe} and \text{BRISQUE} \cite{brisque} for our real-world {\dataName} benchmark for a more robust evaluation.

\subsection{Performance Analysis}

\begin{table}\centering
\caption{Empirical evaluation on {\dataName}.}
\label{tab:metalens_zs}
\vspace{-3mm}
  \renewcommand\arraystretch{1.2}
\resizebox{\columnwidth}{!}{\begin{tabular}{@{}lccccc@{}}\toprule
Method & PSNR $\uparrow$   & SSIM $\uparrow$ & LPIPS $\downarrow$ & NIQE $\downarrow$  & BRISQUE $\downarrow$  \\ \midrule
\textbf{\textit{Zero-shot}} \\
AirNet \cite{li2022all} & 12.69 & 0.341 & 0.62 & 29.57 & 39.96 \\
PromptIR \cite{potlapalli2024promptir} & 13.10 & 0.358 & 0.616 & 29.53 & 34.45 \\
AutoDIR \cite{jiang2023autodir} & 12.65 & 0.255 & 0.654 & 22.45 & 61.29 \\
DiffUIR \cite{zheng2024selective} & 16.97& 0.434 & 0.472 & 12.78 & 47.98 \\
DA-CLIP \cite{luo2023controlling} & 14.41 & 0.312 & 0.687 & 22.21 & 49.16 \\
\rowcolor{YellowGreen!40}
\textbf{\textsc{UniCoRN}} & \textbf{27.93} & \textbf{0.436} & \textbf{0.554} & \textbf{5.27} & \textbf{31.66} \\ \midrule
\textbf{\textit{Fine-tuned}} \\
AirNet \cite{li2022all} & 10.88 & 0.211 & 0.765 & 24.29 & 71.29   \\
MultiWNet \cite{yanny2022deep} & 26.05 & 0.402 &  0.554 & 10.19 & 60.75 \\
\rowcolor{YellowGreen!40}
\textbf{\textsc{UniCoRN}} & \textbf{28.54} & \textbf{0.517} & \textbf{0.434} & \textbf{2.85} & \textbf{27.09} \\
\bottomrule
    \end{tabular}}
\end{table}

\keypoint{Evaluation on {\dataName}:} First, we evaluate on our proposed {\dataName} benchmark on both \textbf{\textit{zero-shot}} and \textbf{\textit{fine-tuned}} settings. While zero-shot evaluation is done on all $1200$ images, in fine-tuned setting we use $800$ images for training and assess on the $400$ image test set (described previously). We compare against recent SoTA approaches for ``universal'' image restoration \cite{zheng2024selective, luo2023controlling, li2022all, jiang2023autodir, potlapalli2024promptir}, \textit{leveraging their open-sourced codes and unified restoration checkpoints.} The empirical results are tabulated in \autoref{tab:metalens_zs}. We consistently outperform all all-in-one methods evaluated in the \textit{unseen} scenario by large margins. For fine-tuning experiments, we observe gains and outperform the end-to-end optical method MultiWNet \cite{yanny2022deep} by a significant margin. Qualitative samples are shown in \autoref{fig:metalens_zeroshot}. Considering \dataName{} is a real-world compound degradation dataset, prior all-in-one approaches fail, while our proposed approach shows robust restoration performance, driving towards truly universal restoration models.

\keypoint{Mixed degradation removal:} We next evaluate {\modelName}'s performance on synthetically generated mixed degradation images, where multiple corruptions are simultaneously present. \cref{tab:synth_degrade} presents the empirical comparisons against prior generalist methods for these tasks. Similar to above, \textit{these experiments also use open-source code and all-in-one restoration checkpoints for prior SoTA models} \cite{jiang2023autodir, li2022all, luo2023controlling}. We observe that {\modelName} surpasses all existing unified restoration works \cite{jiang2023autodir, zheng2024selective, li2022all} by significant margins across all metrics. Furthermore, the poor performance of PromptIR \cite{potlapalli2024promptir} as compared to other approaches shows the prowess of diffusion generative prior for image restoration tasks. \textit{Similar trends were observed in} \cite{zheng2024selective}, \textit{which support our claims}. We remark that these results highly encourage our endeavor to develop a universal image restoration model.

\begin{table}
  \caption{Empirical results on \textbf{mixed degradation} datasets.}
  \vspace{-3mm}
  \label{tab:synth_degrade}
  \centering
  \renewcommand\arraystretch{1.25} %
  \setlength\tabcolsep{2pt} %
  \resizebox{\columnwidth}{!}{
  \begin{tabular}{c|ccc|ccc|ccc}
    \toprule
    \multirow{2}{*}{\textbf{Method}} & \multicolumn{3}{c|}{\textbf{\textit{Blur + Haze}} \cite{reside}} & \multicolumn{3}{c|}{\textbf{\textit{Noise + Blur}}: Div2k \cite{div2k}} & \multicolumn{3}{c}{\textbf{\textit{Low-light + Blur}} \cite{zhou2022lednet}} \\ 
     & PSNR $\uparrow$ & SSIM $\uparrow$ & LPIPS $\downarrow$ & PSNR $\uparrow$ & SSIM $\uparrow$ & LPIPS $\downarrow$ & PSNR $\uparrow$ & SSIM $\uparrow$ & LPIPS $\downarrow$ \\ 
    \midrule
    AutoDIR \cite{jiang2023autodir} & 12.97 & 0.394 & 0.577 & 17.97 & 0.475 & 0.441 & 18.32 & 0.662 & 0.304 \\
    AirNet \cite{li2022all} & 16.99 & 0.480 & 0.569 & 18.56 & 0.564 & 0.401 & 10.20 & 0.130 & 0.515 \\
    PromptIR \cite{potlapalli2024promptir} & 14.75 & 0.431 & 0.528 & 23.30 & 0.601 & 0.455 & 10.24 & 0.096 & 0.501 \\
    DA-CLIP \cite{luo2023controlling} & 14.37 & 0.493 & 0.638 & 21.52 & 0.671 & 0.327 & 16.44 & 0.688 & 0.242 \\
    \rowcolor{YellowGreen!40}
    \textbf{\textsc{UniCoRN}} & \textbf{28.83} & \textbf{0.673} & \textbf{0.212} & \textbf{28.55} & \textbf{0.717} & \textbf{0.162} & \textbf{28.47} & \textbf{0.777} & \textbf{0.149} \\
    \bottomrule
  \end{tabular}
  }
\end{table}

\subsection{Ablation on {\modelName} Design}\label{sec:abla}

We investigate several key components of {\modelName} to justify their role in the overall model design. Our findings (fine-tuned on \dataName{}) are summarised in \autoref{tab:abal}.

\begin{enumerate}[label=(\textbf{\roman*}), leftmargin=*]
\setlength{\itemindent}{0em}
    \item \textbf{Role of task prompt}: In the absence of task prompt the model is left to ``\textit{guess}'' the control type from the task-adaptive mixture-of-experts block. As we see in \cref{tab:abal}, the PSNR drops only by $1.6dB$ proving its generalization capabilities in \textit{unseen} scenarios. 
    \item \textbf{Secondary guidance:} Low level cues help increase the model's performance in the cases of mixed degradations through the use of cheap prior methods.
    \item \textbf{Effectiveness of MLCN \& NAFBlock:} We observe that one of our most impactful additions to the ControlNet architecture is using the MLCN instead of a linear encoder chain and using a NAFBlock \cite{chen2022nafnet} instead of the simple convolutional encoder used in the original architecture. We record an increase in all 3 metrics PSNR, SSIM and LPIPS when using this combination.
    \item \textbf{Curriculum training process:} Finally, the curriculum learning schedule also provides a notable boost in terms of SSIM due to its stable training process.
\end{enumerate}

\begin{table}[!htbp]
\renewcommand{\arraystretch}{1}
\setlength{\tabcolsep}{7pt}
\footnotesize
\centering
\caption{Ablation study on model design.}
\vspace{-0.3cm}
\label{tab:abal}
\begin{tabular}{lccc}
\toprule
\multicolumn{1}{c}{\multirow{1}{*}{Methods}} & \multirow{1}{*}{PSNR~$\uparrow$}  & \multirow{1}{*}{SSIM~$\uparrow$}  & \multirow{1}{*}{LPIPS~$\downarrow$}\\
\cmidrule(lr){1-4}
w/o task prompt & 27.12 & 0.670 & 0.214 \\
w/o secondary guidance  & 27.45 & 0.624 & 0.216 \\
w/o MLCN+NAFBlock       & 26.55 & 0.510 & 0.293 \\
w/o NAFBlock    & 27.57 & 0.668 & 0.242 \\
w/o Curriculum Learning     & 27.89 & 0.660 & 0.190 \\ \cmidrule(lr){1-4}
\rowcolor{YellowGreen!40}
\textbf{{\modelName} \textit{(full)}}       & \bf 28.74 & \bf 0.772 & \bf 0.173 \\ \bottomrule
\end{tabular}
\vspace{-0.2cm}
\end{table}

\begin{table*}[tbp]
  \caption{Empirical results and SoTA comparison across various single-degradation restoration tasks.}
  \vspace{-3mm} 
  \label{tab:single_degrade}
  \centering
  \renewcommand\arraystretch{1.1} %
  \setlength\tabcolsep{5pt} %
  \resizebox{\textwidth}{!}{
  \begin{tabular}{cccccccccccc}
    \toprule
    \multicolumn{3}{c}{\textbf{\textit{Dehazing:}} RESIDE \cite{reside}} & \multicolumn{3}{c}{\textbf{\textit{Motion Deblurring}}: GoPro \cite{gopro}} & \multicolumn{3}{c}{\textbf{\textit{Low-Light Enhancement:}} LOL \cite{wei2018deep}} & \multicolumn{3}{c}{\textbf{\textit{Denoising}}: CBSD68 \cite{cbsd68}} \\ 
    \cmidrule{1-12}
    \textbf{Method} & PSNR $\uparrow$ & LPIPS $\downarrow$ & \textbf{Method} & PSNR $\uparrow$ & LPIPS $\downarrow$ & \textbf{Method} & PSNR $\uparrow$ & LPIPS $\downarrow$ & \textbf{Method} & PSNR $\uparrow$ & LPIPS $\downarrow$ \\ 
    \midrule
    \textbf{\textit{Non-diffusion approaches}} \\
    AirNet \cite{li2022all} & 23.70 & 0.073 & AirNet \cite{li2022all} & 28.31 & \underline{0.122} & AirNet \cite{li2022all} & 19.69 & 0.151 & AirNet \cite{li2022all} & 26.13 & 0.394 \\
    PromptIR \cite{potlapalli2024promptir} & \textbf{33.11} & \underline{0.030} & PromptIR \cite{potlapalli2024promptir} & \textbf{31.02} & 0.131 & PromptIR \cite{potlapalli2024promptir} & 21.23 & 0.145 & PromptIR \cite{potlapalli2024promptir} & 28.82 & \underline{0.170} \\
    \textbf{\textit{Diffusion-based approaches}} \\
    AutoDIR \cite{jiang2023autodir} & 20.55 & \textbf{0.025} & AutoDIR \cite{jiang2023autodir} & 28.02 & 0.167 & AutoDIR \cite{jiang2023autodir} & 20.98 & 0.160 & AutoDIR \cite{jiang2023autodir} & \underline{29.24} & 0.267 \\
    DiffUIR-L \cite{zheng2024selective} & 21.32 & 0.038 & DiffUIR-L \cite{zheng2024selective} & 25.12 & \textbf{0.035} & DiffUIR-L \cite{zheng2024selective} & \underline{25.81} & \textbf{0.040} & IR-SDE \cite{irsde} & 28.09 & \textbf{0.103} \\
    DA-CLIP \cite{luo2023controlling} & 23.77 & 0.083 & DA-CLIP \cite{luo2023controlling} & 27.12 & 0.134 & DA-CLIP \cite{luo2023controlling} & 22.09 & \underline{0.114} & DA-CLIP \cite{luo2023controlling} & 24.36 & 0.272 \\ \midrule
    \rowcolor{YellowGreen!40}
    \textbf{\textsc{UniCoRN}} & \underline{29.90} & 0.174 & \textbf{\textsc{UniCoRN}} & \underline{28.90} & 0.155 & \textbf{\textsc{UniCoRN}} & \textbf{28.30} & 0.159 & \textbf{\textsc{UniCoRN}} & \textbf{29.30} & 0.301 \\
    \bottomrule
  \end{tabular}
  }
\end{table*}

\begin{figure*}[!htbp]
	\scriptsize
	\centering
	\newcommand{\h}{0.105}
	\newcommand{\wa}{0.12}
	\newcommand{\wb}{0.16}
	\newcommand{\g}{-0.7mm}

 	\setlength\tabcolsep{1.8pt}
	\renewcommand{\arraystretch}{1}
	\resizebox{1.00\linewidth}{!} {
			\renewcommand{\h}{0.15}
			\newcommand{\w}{0.200}
				\begin{adjustbox}{valign=t}
					\begin{tabular}{cccccc}
						\includegraphics[height=\h \textwidth, width=\w \textwidth]{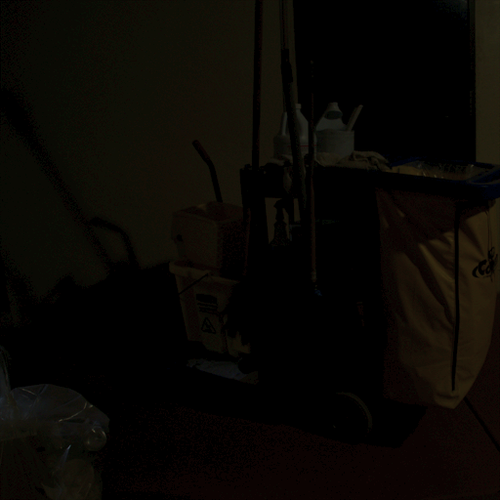} \hspace{\g} &
						\includegraphics[height=\h \textwidth, width=\w \textwidth]{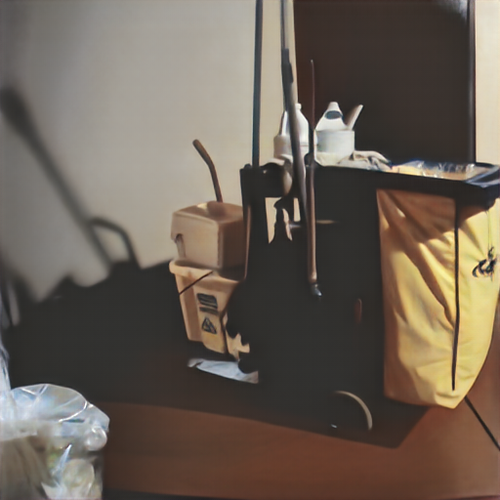} \hspace{\g} &
                        \includegraphics[height=\h \textwidth, width=\w \textwidth]{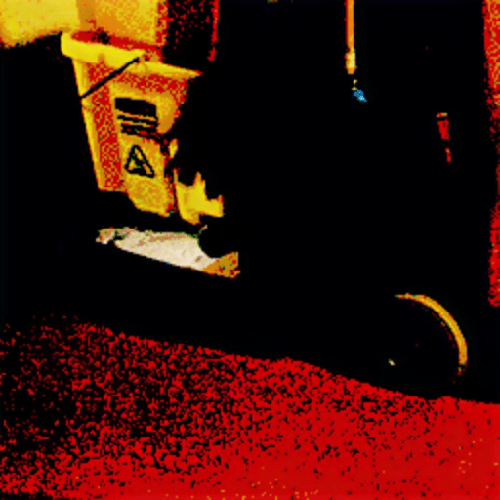} \hspace{\g} &
                        \includegraphics[height=\h \textwidth, width=\w \textwidth]{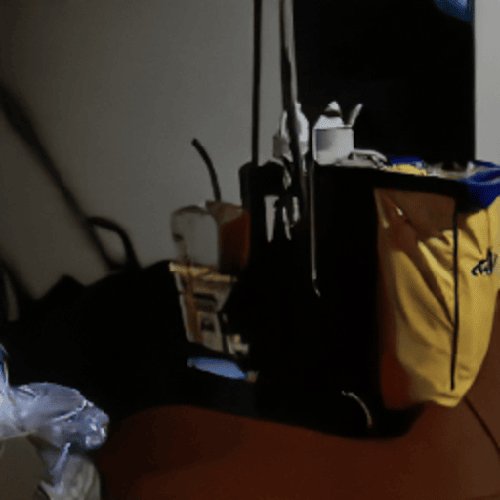} \hspace{\g} &
                        \includegraphics[height=\h \textwidth, width=\w \textwidth]{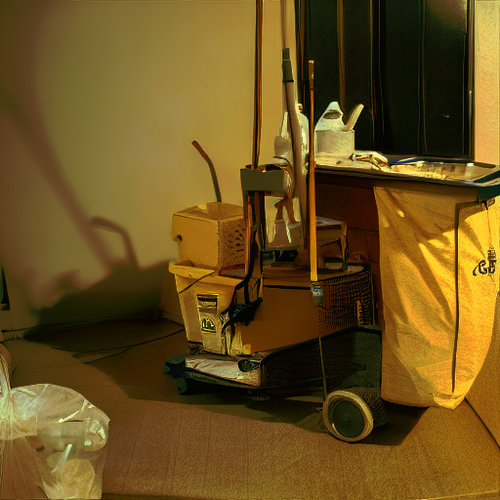} \hspace{\g} &
                        \includegraphics[height=\h \textwidth, width=\w \textwidth]{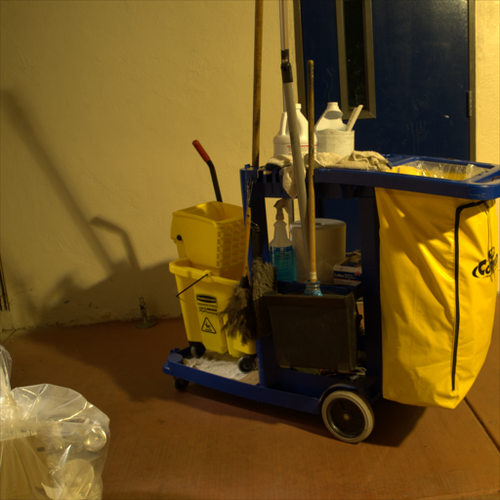}
                        						\\
						Low Light Input \hspace{\g} &
                        AutoDIR~\cite{jiang2023autodir}  \hspace{\g} & 
                        DiffUIR-L~\cite{zheng2024selective}  \hspace{\g} & 
                        DA-CLIP~\cite{luo2023controlling}  \hspace{\g} & 
                        {\modelName} (Ours)  \hspace{\g} & 
                        GT
                                    \\
						\includegraphics[height=\h \textwidth, width=\w \textwidth]{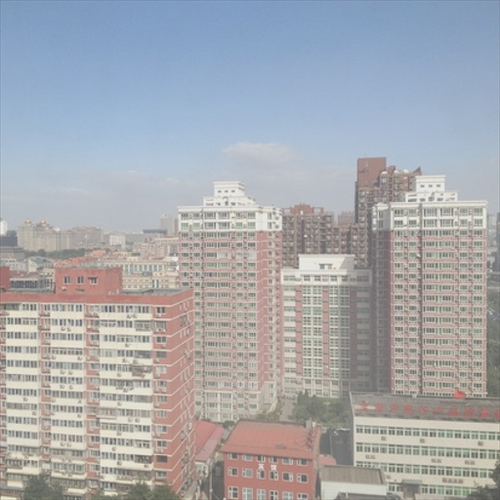} \hspace{\g} &
						\includegraphics[height=\h \textwidth, width=\w \textwidth]{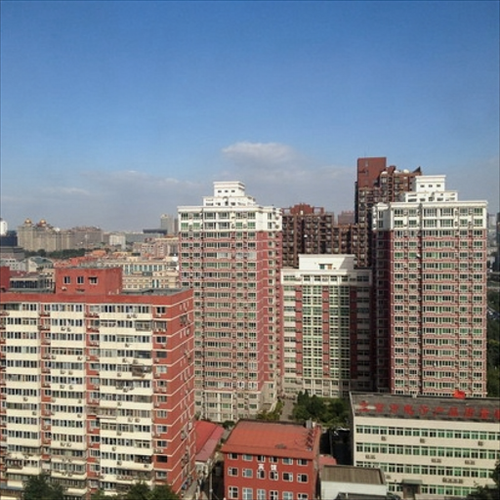} \hspace{\g} &
						\includegraphics[height=\h \textwidth, width=\w \textwidth]{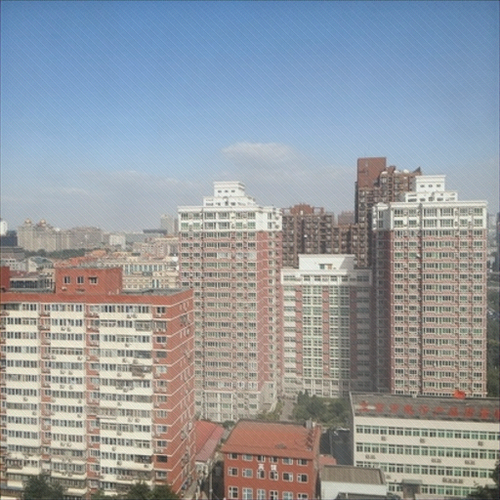} \hspace{\g} &
                        \includegraphics[height=\h \textwidth, width=\w \textwidth]{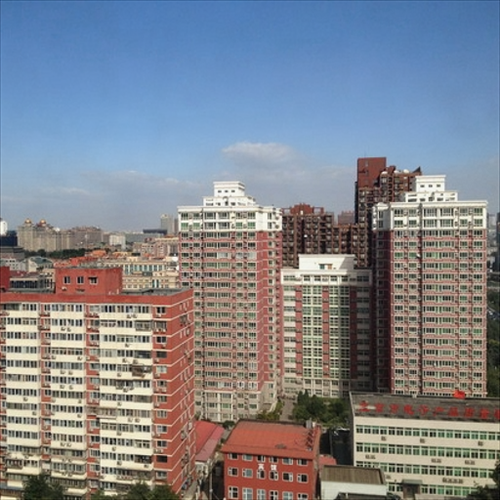} \hspace{\g} &
                        \includegraphics[height=\h \textwidth, width=\w \textwidth]{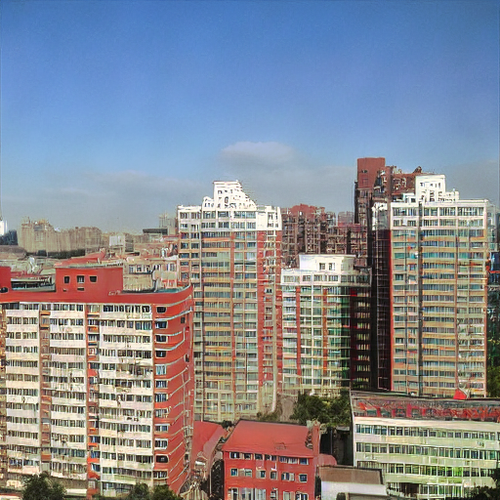} \hspace{\g} &
                        \includegraphics[height=\h \textwidth, width=\w \textwidth]{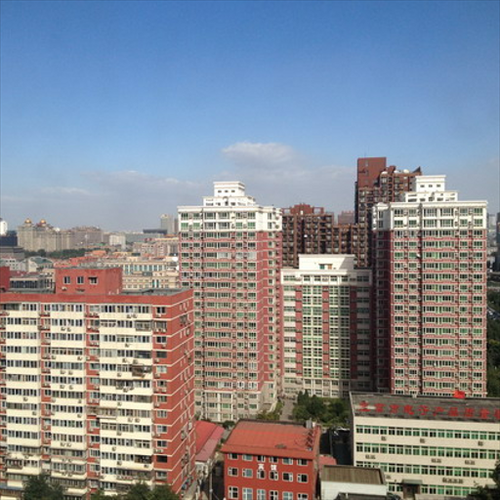} 
						\\
						Hazy Input \hspace{\g} &
						AutoDIR~\cite{jiang2023autodir} \hspace{\g} & 
                        Airnet~\cite{li2022all} \hspace{\g} & 
                        PromptIR~\cite{potlapalli2024promptir} \hspace{\g} & 
                        {\modelName} (Ours) \hspace{\g} & 
                        GT 
                                                \\
						\includegraphics[height=\h \textwidth, width=\w \textwidth]{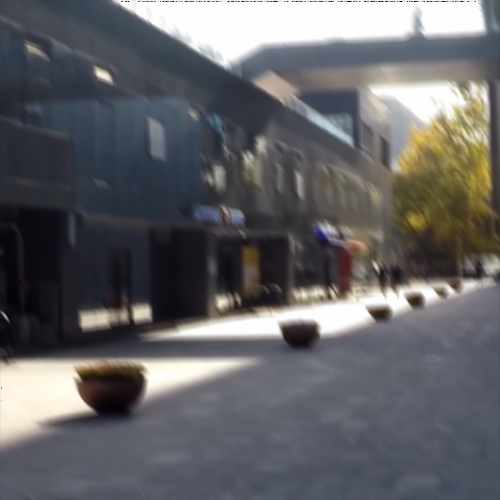} \hspace{\g} &
						\includegraphics[height=\h \textwidth, width=\w \textwidth]{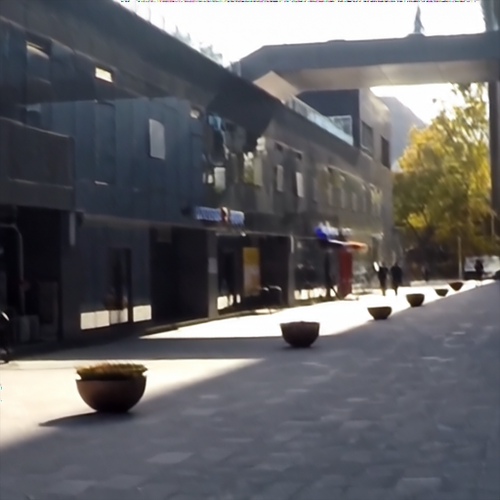} \hspace{\g} &
						\includegraphics[height=\h \textwidth, width=\w \textwidth]{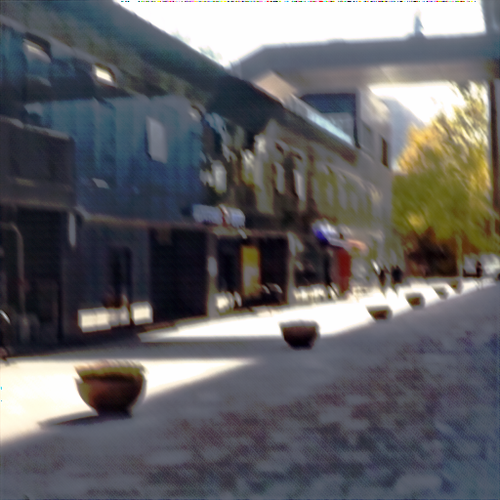} \hspace{\g} &
                        \includegraphics[height=\h \textwidth, width=\w \textwidth]{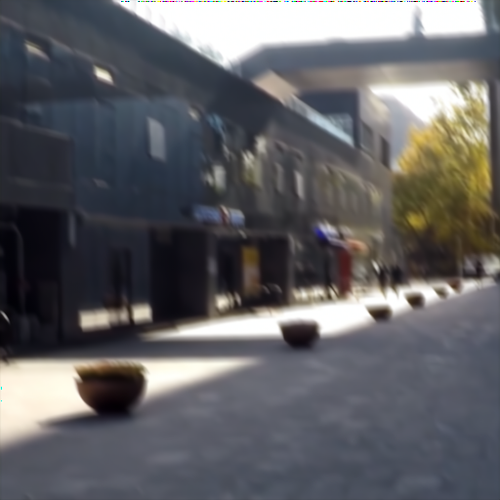} \hspace{\g} &
                        \includegraphics[height=\h \textwidth, width=\w \textwidth]{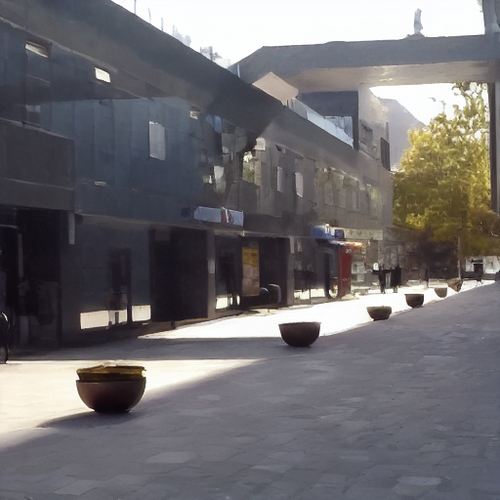} \hspace{\g} &
                        \includegraphics[height=\h \textwidth, width=\w \textwidth]{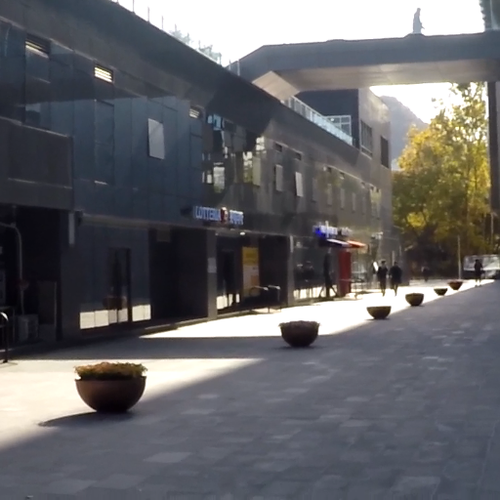} 
						\\
						Blurry Input \hspace{\g} &
						AutoDIR~\cite{jiang2023autodir} \hspace{\g} & 
                        Airnet~\cite{li2022all} \hspace{\g} & 
                        PromptIR~\cite{potlapalli2024promptir} \hspace{\g} & 
                        {\modelName} (Ours) \hspace{\g} & 
                        GT 

                           \\
						\includegraphics[height=\h \textwidth, width=\w \textwidth]{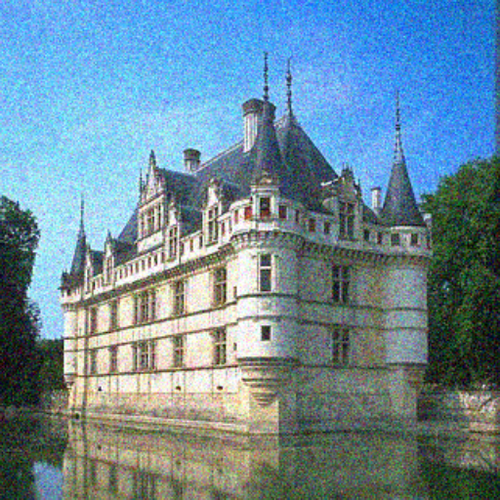} \hspace{\g} &
						\includegraphics[height=\h \textwidth, width=\w \textwidth]{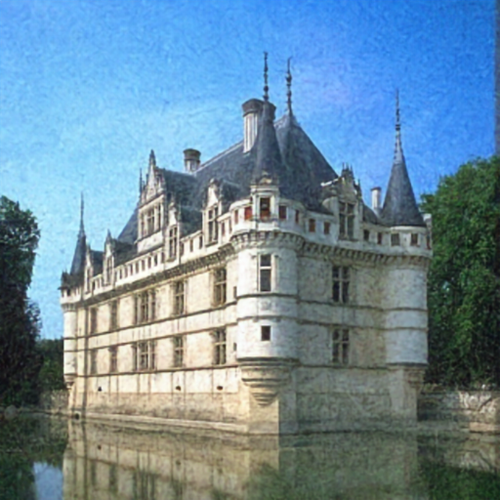
                        } \hspace{\g} &
						\includegraphics[height=\h \textwidth, width=\w \textwidth]{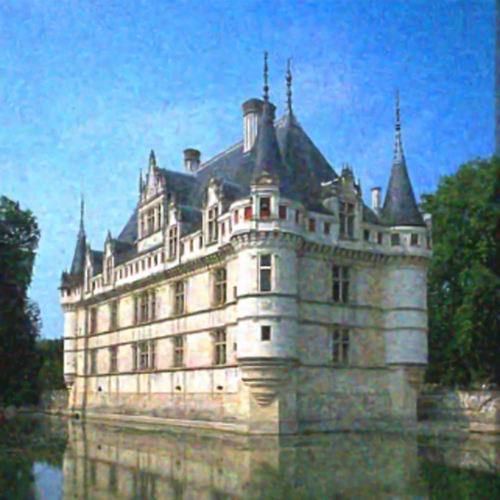} \hspace{\g} &
                        \includegraphics[height=\h \textwidth, width=\w \textwidth]{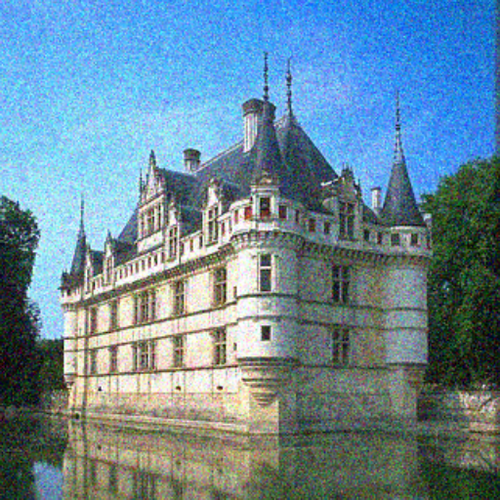} \hspace{\g} &
                        \includegraphics[height=\h \textwidth, width=\w \textwidth]{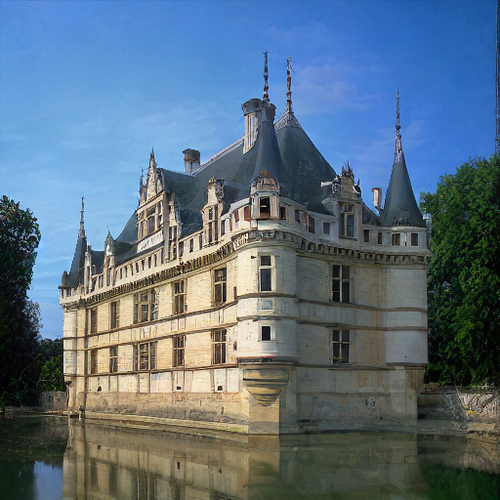} \hspace{\g} &
                        \includegraphics[height=\h \textwidth, width=\w \textwidth]{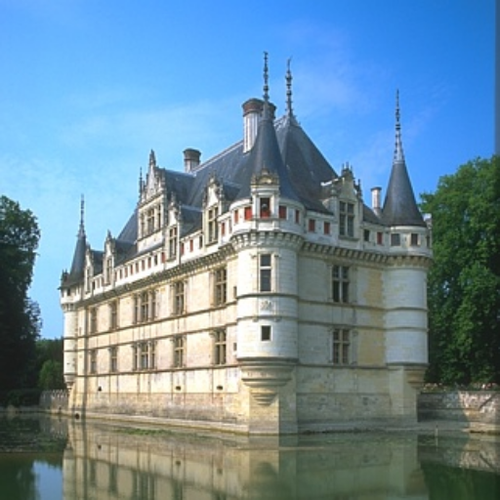} 
						\\
						Noisy Input \hspace{\g} &
						AutoDIR~\cite{jiang2023autodir} \hspace{\g} & 
                        Airnet~\cite{li2022all} \hspace{\g} & 
                        PromptIR~\cite{potlapalli2024promptir} \hspace{\g} & 
                        {\modelName} (Ours) \hspace{\g} & 
                        GT 

					\end{tabular}
				\end{adjustbox}
		 }

 \vspace{-3mm}
	\caption{Qualitative results on single-degradation restoration tasks. Best viewed if zoomed in. }
	\label{fig:single}
 \vspace{-3mm}
\end{figure*}

\subsection{Single-degradation removal}

Although not our primary objective, we also evaluate {\modelName} on several single-degradation restoration tasks: dehazing \cite{reside}, motion deblurring \cite{gopro}, relighting \cite{wei2018deep} and denoising \cite{cbsd68}. We compare \modelName{'s} performance against other generalist restoration models \cite{li2022all, potlapalli2024promptir, jiang2023autodir, zheng2024selective, luo2023controlling}; the results are presented in \autoref{tab:single_degrade}. {\modelName} achieves the best PSNR score for relighting and denoising tasks, while being the second-best performer for dehazing and motion deblurring, surpassed by PromptIR. However, it should be noted that PromptIR \cite{potlapalli2024promptir} has \textit{degradation-specific optimized checkpoints}; although they struggle for relighting and denoising tasks. Among diffusion-based approaches, {\modelName} emerges as the best model across all tasks in terms of PSNR values. The highly competitive empirical values can be further supported through visual results shown in \autoref{fig:single}, which shows our model restores even extreme degradations.

\subsection{Further analysis}

\keypoint{Choice of MLCN block:}We further justify the use of NAFBlock \cite{chen2022nafnet} by comparing with channel attention \cite{mprnet} and residual block \cite{li2020all} in \autoref{tab:abal_encoder}, observing that NAFBlock dominates in terms of faster convergence and higher perceptual image quality when used at the control encoder head.

\begin{table}[!htbp]
\renewcommand{\arraystretch}{1}
\setlength{\tabcolsep}{7pt}
\footnotesize
\centering
\caption{Ablation study on MLCN design.}
\vspace{-0.3cm}
\label{tab:abal_encoder}
\begin{tabular}{ccc}
\toprule
\multicolumn{1}{c}{\multirow{1}{*}{Encoder choice}} & \multirow{1}{*}{Time to converge (hr)}  
 & \multirow{1}{*}{LPIPS~$\downarrow$}\\
\cmidrule(lr){1-3}
{Basic encoder}       &  172 &  0.293 \\ 
$+$ ResBlock \cite{li2020all} & 144  & 0.242 \\
$+$ CABlock  \cite{mprnet}  & 120 & 0.181 \\
 \rowcolor{YellowGreen!40}
$+$ \textbf{NAFBlock} \cite{chen2022nafnet}   & \textbf{120}     &  \textbf{0.173} \\
\bottomrule
\end{tabular}
\end{table}

\keypoint{Effect of batch size:}While \cite{controlnet} observed that higher batch sizes converged faster than training for longer iterations, in \autoref{tab:abal_batch} we find that an optimal batch size of $8$ for relatively small-sized dataset ($\sim$ 20K) reaches good restoration quality.

\begin{table}[!htbp]
\renewcommand{\arraystretch}{1}
\setlength{\tabcolsep}{7pt}
\footnotesize
\centering
\caption{Ablation study on optimal batch size.}
\vspace{-0.3cm}
\label{tab:abal_batch}
\resizebox{\columnwidth}{!}{
\begin{tabular}{ccccc}
\toprule
\multicolumn{1}{c}{\multirow{1}{*}{Batch size}} & \multirow{1}{*}{Time to converge (hr)}  & \multirow{1}{*}{PSNR~$\uparrow$}  & \multirow{1}{*}{SSIM~$\uparrow$}  & \multirow{1}{*}{LPIPS~$\downarrow$}\\
\cmidrule(lr){1-5}
1        & 160 & 25.45 & 0.471 & 0.383 \\
\textbf{8}   & \textbf{120} & \textbf{28.74} & \textbf{0.772} & \textbf{0.173} \\
32   & 192 & 26.29 & 0.601 & 0.215 \\
128   & 210 & 26.13 & 0.551& 0.243 \\ 
\bottomrule
\end{tabular}}
\end{table}

\keypoint{Inference time comparison:}We compare the inference times of various restoration models in \autoref{tab:inference_speeds} and observe that \modelName{} is $3-5\times$ faster than diffusion-based approaches \cite{jiang2023autodir, zheng2024selective, luo2023controlling}, almost comparable to the non-diffusion models PromptIR \cite{potlapalli2024promptir} and AirNet \cite{li2022all}. Thus, our approach delivers a robust, efficient universal restoration model, surpassing prior SoTA methods.

\begin{table}[!htbp]
    \centering
    \caption{Inference times of various restoration methods.}
    \label{tab:inference_speeds}
    \vspace{-0.3cm}
    \renewcommand{\arraystretch}{1.2} %
    \setlength{\tabcolsep}{2pt} %
    \resizebox{\columnwidth}{!}{
    \begin{tabular}{ccccccc}
        \toprule
        \textbf{Model} & {AutoDIR} & {DA-CLIP} & {DiffUIR}  & {AirNet} & {PromptIR} & {\cellcolor{YellowGreen!40}\textbf{\modelName{}}} \\ \midrule
        \textbf{Inference time (s)} & 35 & 23 & 20 & 7 & 4 & {\cellcolor{YellowGreen!40}7} \\ 
        \bottomrule
    \end{tabular}}
    \vspace{-3mm}
\end{table}

\def\modelName{\textsc{UniCoRN}}
\def\dataName{\textsc{MetaRestore}}

\section{Conclusion}
\vspace{-1mm}

In this paper, we presented \modelName, a unified image restoration approach designed to handle multiple real-world degradations in images. 
Our approach circumvents the need for prior knowledge of type and level of degradations by leveraging low-level image features to manage unknown degradation types. 
We also introduced \dataName~dataset of metalens-degraded images. With experiments across a variety of challenging restoration benchmarks, we demonstrate that \modelName~achieves significant performance improvements. Optimizing diffusion models and mitigating hallucinations can enhance unified degradation removal for real-world applications.

{
    \small
    \bibliographystyle{ieeenat_fullname}
    \bibliography{main}
}

\end{document}